\documentclass[nonatbib]{article}

 \usepackage[main, final]{neurips_2026}

\usepackage[utf8]{inputenc} 
\usepackage[T1]{fontenc}    
\usepackage{hyperref}       
\usepackage{url}            
\usepackage{booktabs}       
\usepackage{amsfonts}       
\usepackage{nicefrac}       
\usepackage{microtype}      
\usepackage[dvipsnames]{xcolor}       
\usepackage{graphicx} 
\usepackage{amsmath, amsthm}
\usepackage[capitalize]{cleveref}
\newtheorem{theorem}{Theorem}
\newtheorem{proposition}{Proposition}
\usepackage{etoc}
\usepackage[numbers, sort&compress]{natbib}
\usepackage{wrapfig}
\usepackage{algorithm}
\usepackage{algorithmic}
\usepackage{pifont}   
\usepackage{subcaption}

\newcommand{\cmark}{\textcolor{green!80!black}{\ding{51}}}
\newcommand{\xmark}{\textcolor{red}{\ding{55}}}          

\title{Flow Matching Policy Optimization with Mirror Descent and Entropy Constraints}

\author{%
  Ting Gao, Stavros Orfanoudakis, Nan Lin, Winnie Daamen, Serge Hoogendoorn, Elvin Isufi\\
  Delft University of Technology\\
  \texttt{\{T.Gao-1, S.Orfanoudakis, N.Lin, W.Daamen, S.P.Hoogendoorn, E.Isufi-1\}@tudelft.nl}\\
}
\begin{document}

\maketitle

\begin{abstract}
Balancing policy expressiveness with the exploration-exploitation trade-off is a core challenge in online Reinforcement Learning (RL). While Stochastic Differential Equation (SDE)-based diffusion policies can represent complex, multimodal action distributions, they suffer from two critical limitations: their stochastic reverse processes render entropy intractable (necessitating heuristic exploration), and computing policy gradients through long denoising chains is expensive and unstable. In this work, we show that ODE-based flow matching inherently resolves these issues by enabling both simulation-free policy optimization and tractable entropy computation. Building on this, we introduce Flow Matching Policy Optimization with Mirror Descent and Entropy Constraints (FMER). Our framework exploits this insight in three ways. First, we derive an analytic entropy objective that corrects for the density distortion caused by the $\tanh$ transformation (mapping an unbounded latent space to bounded actions), thereby facilitating entropy-constrained policy optimization. Second, we theoretically establish that minimizing an advantage-weighted conditional flow matching loss acts as a simulation-free surrogate for policy mirror descent. This steers the velocity field toward high-value regions while entirely avoiding backpropagation through the ODE solver. Finally, we dynamically tune the mirror descent temperature based on the effective sample size to enforce a robust trust region during training. Empirical evaluations demonstrate that FMER achieves superior performance on the challenging sparse-reward FrankaKitchen environment, while maintaining competitive results across standard dense-reward MuJoCo benchmarks. Code is available at \url{https://github.com/gtanaelle/FMER.git}.
\end{abstract}

\addtocontents{toc}{\protect\setcounter{tocdepth}{0}}
\section{Introduction}
In continuous control RL, standard unimodal Gaussian policies often fail to capture complex action landscapes. Many tasks exhibit multiple distinct, state-dependent high-return behaviors, necessitating policies capable of representing multi-peaked distributions~\cite{haarnoja2017reinforcement}. This has motivated the development of generative policy classes that can represent rich, multimodal action distributions by transforming noise samples into actions conditioned on the state. Diffusion policies transform this via iterative denoising~\cite{dacer,dipo}, often relying on simulating a reverse-time SDE. Meanwhile, Flow Matching (FM) formulations~\citep{DBLP:conf/iclr/LipmanCBNL23,fmpo} provide an ODE-based alternative that learns the underlying transport by directly regressing a vector field, closely related to Continuous Normalizing Flows~\citep{chen2018neural}.

Diffusion policies have first gained prominence in offline RL~\citep{DBLP:conf/iclr/WangHZ23, decision_diffuser}, typically formulated as behavior cloning of expert data. However, extending these models to online RL is non-trivial due to the absence of pre-collected expert data. Existing online approaches fall into two categories: (1) using the diffusion model purely as an approximation tool and backpropagating policy gradients through the sampling chain \cite{dacer}, a process that is computationally expensive and susceptible to high-variance gradients \cite{ren2024diffusionpolicypolicyoptimization, ding2024diffusion}; (2) leveraging the generative capability to iteratively improve the policy by regressing towards high-value samples selected by a critic \cite{dipo, ding2024diffusion, ma2025efficientonlinereinforcementlearning}. Our work follows the second paradigm.

While SDE frameworks excel at approximating complex distributions, they face a fundamental limitation in online RL: their stochastic reverse process renders policy entropy intractable, yet without tractable entropy, maximum entropy exploration requires crude heuristics (e.g., uniform distribution assumptions~\cite{ding2024diffusion}). Removing stochasticity by switching to ODE-based methods recovers tractable entropy but backpropagating through the ODE integration is numerically unstable~\cite{ding2024diffusion, zhang2026sac}.

Recently, several ODE-based FM policies have emerged. For instance, \cite{zhang2026sac} address gradient instability using sequence modeling techniques, while \cite{li2026boostingmaximumentropyreinforcement} achieve one-step FM generation within a maximum entropy framework. However, these approaches overlook the policy mirror descent perspective, which imposes crucial behavioral constraints to prevent policy collapse in extremely sparse environments or under perturbed critic estimations. Additionally, in their pursuit of reducing integration steps, these prior methods compromise on closed-form entropy by relying on entropy surrogates or augmented ODEs. 

As shown in Table~\ref{tab:concept_comparison}, existing methods trade off at least one of three desirable properties: tractable entropy, simulation-free training, and a principled mirror descent trust region. FMER is designed to satisfy all three simultaneously. We propose Flow Matching Policy Optimization with Mirror Descent and Entropy Constraints (FMER), which resolves these challenges by adopting the Optimal Transport (OT) conditional FM objective~\citep{DBLP:conf/iclr/LipmanCBNL23} to preserve simulation-free training, deriving a closed-form entropy expression that corrects for the $\tanh$ density distortion in bounded action spaces, and incorporating a mirror descent trust region via advantage-weighted updates. Each design choice directly addresses one of the three challenges above. While FMER shares the ideal policy mirror descent target with DPMD~\cite{ma2025efficientonlinereinforcementlearning}, it realizes this target through simulation-free weighted conditional FM rather than reweighted denoising score matching.

Our contributions are threefold:
\begin{itemize}
    \item \textbf{FM Policy Optimization with Mirror Descent and Entropy Constraints.} We propose FMER, which leverages an ODE-based Flow Matching policy to improve expressiveness. FMER incorporates an explicit entropy constraint that directly encourages high-entropy policies, thereby improving exploration. In addition, FMER adopts the standard policy mirror descent target (shared with DPMD \cite{ma2025efficientonlinereinforcementlearning}) and contributes a simulation-free realization of it with self-normalized, ESS-adaptive weights.

    \item \textbf{Theoretical Foundations.} We derive a tractable entropy objective with an unbiased divergence estimator, which accounts for the density distortion induced by $\tanh$ mapping from the unbounded FM latent space to a bounded action space. This enables closed-form entropy-constrained optimization without surrogates or augmented ODEs. Under regularity and realizability assumptions, we further prove that any global minimizer of the weighted conditional FM loss induces the ideal policy mirror descent target, enabling efficient simulation-free training. 

    \item \textbf{Empirical Validation.} We evaluate FMER against eight baselines spanning standard RL, diffusion-based RL, and FM-based RL methods. Experiments are conducted on 2D multi-goal environments, multi-task long-horizon and sparse-reward FrankaKitchen tasks, and dense-reward MuJoCo benchmarks. Together, these experiments and ablation studies validate that policy multimodality, principled entropy control, and mirror descent trust regions each play a necessary role in FMER's performance.
\end{itemize}

\begin{table}[!t]
    \centering
    \caption{Conceptual comparison between FMER and diffusion- and flow-based baselines.}
    \label{tab:concept_comparison}
    \resizebox{\linewidth}{!}{%
    \begin{tabular}{lcccc}
        \toprule
        \textbf{Method} & \textbf{Policy Mirror Descent} & \textbf{Entropy Constrained} & \textbf{Unbiased Entropy Objective}& \textbf{Simulation-Free Training} \\
        \midrule
        DIPO~\citep{dipo}                                        & \xmark & \xmark & \xmark & \cmark \\
        DPMD~\citep{ma2025efficientonlinereinforcementlearning}  & \cmark & \xmark & \xmark & \cmark \\
        QVPO~\citep{ding2024diffusion}                           & \xmark & \cmark & \xmark & \cmark \\
        SAC-Flow~\citep{zhang2026sac}                            & \xmark & \cmark & \xmark & \xmark \\
        FLAME~\citep{li2026boostingmaximumentropyreinforcement}   & \xmark & \cmark &\xmark  & \xmark  \\
        \textbf{FMER (Ours)}                                     & \cmark & \cmark & \cmark & \cmark \\
        \bottomrule
    \end{tabular}}
\end{table}

\section{Related work}
\noindent\textbf{Diffusion policies.} Diffusion policies parameterize the policy network as a diffusion model and first gained prominence in offline RL \cite{DBLP:conf/iclr/WangHZ23, kang2023efficient, DBLP:conf/iclr/Chen0Y0023, hansen2023idql}. Online extensions include DIPO~\citep{dipo}, which imitates Q-gradient-updated replay actions; QSM~\citep{qsm}, which aligns score estimates with critic gradients; DACER~\citep{dacer} and DIME~\citep{dime}, which backpropagate policy gradients through the diffusion sampling chain; and critic-weighted regression methods such as QVPO~\citep{ding2024diffusion} and DPMD~\citep{ma2025efficientonlinereinforcementlearning}. These approaches face expensive optimization through long sampling chains, and heuristic entropy approximations that bias exploration.

\noindent\textbf{Flow Matching (FM) Policies.}
FM replaces stochastic reverse-time dynamics with a deterministic ODE over a learned vector field \citep{DBLP:conf/iclr/LipmanCBNL23, chen2018neural}, offering simulation-free training and easier-to-control sampling. Prior offline applications \citep{DBLP:conf/iclr/ZhangZG25, DBLP:conf/humanoids/RouxelDCIM25} and online methods including Flow-GRPO \citep{liu2025flowgrpo}, FPO \citep{fmpo}, FPMD \citep{FPMD}, GoRL \citep{zhang2026evolvingdiffusionflowmatching}, and FM-IRL \citep{wan2025fmirlflowmatchingrewardmodeling} have explored this direction. RWFM and GRPO~\citep{rwfm} improve demonstration-initialized action-chunk planners, whereas FMER learns from scratch through online interaction in sequential decision problems. RFM \cite{li2026reverseflowmatchingunified} bridges diffusion policies and flow policies under the MaxEnt principle. Most related to ours are SAC-Flow \cite{zhang2026sac} and FLAME \cite{li2026boostingmaximumentropyreinforcement}, which forgo conditional FM's simulation-free advantage by backpropagating through ODE integration, reducing steps for efficiency at the cost of closed-form entropy: SAC-Flow mitigates gradient instability via recurrent architectures, while FLAME achieves single-step generation but relies on entropy surrogates or augmented ODE dynamics. In contrast, FMER preserves simulation-free training and derives a closed-form entropy expression. See Appendix~\ref{appendix:connection_adv}–\ref{appendix:connection_pmd} for detailed comparisons with prior works.

\section{Preliminaries}
\subsection{Reinforcement learning} 
The environment is modeled as a Markov Decision Process (MDP) defined by the tuple $(\mathcal{S}, \mathcal{A}, p, r, \rho_0, \gamma)$ where $\mathcal{S}$ and $\mathcal{A}$ are the continuous state and action spaces. The transition density $p:\mathcal{S}\times\mathcal{A}\times\mathcal{S}\rightarrow [0, \infty)$ describes the likelihood of transitioning to state $s_{i+1}$ given action $a_i$ in state $s_i$. The bounded reward function $r:\mathcal{S}\times\mathcal{A}\rightarrow\mathbb{R}$ gives a scalar reward after an action is taken. $\rho_0$ denotes the initial state density, and $\gamma \in [0, 1)$ is the discount factor. The RL policy is represented as $\pi(a|s): \mathcal{S}\times \mathcal{A}\rightarrow [0, \infty)$, denoting the probability density of taking action $a_i$ at state $s_i$. Starting from an initial state $s_0$, the agent interacts with the environment to get a trajectory  by following the policy $\pi$. We denote the trajectory distribution induced by the policy as $\mathcal{T} \sim \rho_\pi$.

The value function $V_\pi(s)=\mathbb{E}_{\mathcal{T}\sim\rho_\pi}[\sum_{i=0}^{\infty}\gamma^i r_{i} \mid s_0=s]$ estimates the expected discounted future reward starting from any given state $s_0=s$. Similarly, the action-value function $Q_{\pi}(s,a)=\mathbb{E}_{\mathcal{T}\sim\rho_\pi}[\sum_{i=0}^{\infty}\gamma^i r_{i} \mid s_0=s, a_0=a]$ estimates the expected return after taking action $a$ at state $s$. The advantage function $A_\pi(s, a)=Q_\pi(s, a)-V_\pi(s)$ measures the relative benefit of taking action $a$ over the average action at state $s$. The goal of RL is to maximize the expected cumulative reward $J(\pi)=\mathbb{E}_{\mathcal{T}\sim\rho_\pi}[\sum_{i=0}^{\infty}\gamma^i r_i]$.

\subsection{Flow Matching}
\label{sec:prelim_FM}
Flow Matching~\cite{DBLP:conf/iclr/LipmanCBNL23} is a generative framework that models a probability path $p_t$ transporting a simple noise distribution $p_0$ (at $t=0$) to a complex data distribution $p_{\text{data}}$ (at $t=1$). This transport is governed by a time-dependent ODE:
\begin{equation}
\frac{d}{dt}x_t = v_\theta(x_t, t),
\label{eq:FM-ODE}
\end{equation}
where $v_\theta$ is a neural vector field parameterized by $\theta$. To train this field, it is standard to define a target conditional probability path \cite{DBLP:conf/iclr/LipmanCBNL23}. In this work, we utilize the Optimal Transport (OT) path, which linearly interpolates between noise $x_0 \sim \mathcal{N}(0, \mathbf{I})$ and data $x_1 \sim p_{\text{data}}$:
$x_t = (1-t)x_0 + t x_1.$

This path corresponds to a conditional vector field $u_t^\text{target}(x_t|x_0, x_1)$ that generates the trajectory from $x_0$ to $x_1$. For the Optimal Transport (OT) path, this vector field is the constant velocity:
\begin{equation}
u_t^\text{target}(x_t|x_0, x_1) = \frac{d}{dt}x_t = x_1 - x_0.
\label{eq:target_field}
\end{equation}
Since the true marginal vector field is intractable, \cite{DBLP:conf/iclr/LipmanCBNL23} shows that minimizing the Conditional Flow Matching (CFM) loss is equivalent to minimizing the intractable marginal FM loss (up to a constant):
\begin{equation}
\mathcal{L}_{\text{CFM}}(\theta) = \mathbb{E}_{t, x_1 \sim p_{\text{data}}, x_0 \sim p_0} \left[ \left\| v_\theta(x_t, t) - u_t^\text{target}(x_t|x_0,x_1) \right\|^2 \right].
\label{eq:cfm_loss}
\end{equation}

\section{Method}
\label{sec:optimization_objective}
In this section, we introduce the Flow Matching Policy Optimization with Mirror Descent and Entropy Constraints (FMER) framework. 

To stabilize learning, Policy Mirror Descent \cite{tomar2020mirror, lan2023policy} 
formulates the policy update as a constrained optimization problem that maximizes 
the expected advantage while penalizing the Kullback-Leibler (KL) divergence from 
the current policy $\pi_k$:
\begin{equation}
\pi_{k+1} = \arg\max_{\pi}\mathbb{E}_{s\sim\mathcal{D}}\Big[\mathbb{E}_{a\sim\pi(\cdot|s)}[A^{\pi_k}(s,a)] - \tau D_{\mathrm{KL}}(\pi(\cdot|s)\,\|\,\pi_k(\cdot|s))\Big]
\label{eq:pmd_objective}
\end{equation}
where $\tau$ is the temperature parameter controlling the strength of the KL 
penalty, and $\mathcal{D}$ is the replay buffer. Following the same reasoning 
as AWR \cite{peng2019advantage}, denoting $Z(s)=\mathbb{E}_{a\sim\pi_k(\cdot|s)}\left[\exp\left(\frac{A^{\pi_k}(s,a)}{\tau}\right)\right]$, Eq.~\eqref{eq:pmd_objective} yields a 
closed-form objective (proof in Appendix \ref{appendix:pmd_proof}):
\begin{equation}
    \max_\theta \mathbb{E}_{s \sim \mathcal{D}, a\sim \pi_k(\cdot|s)} 
    \left[ \frac{1}{Z(s)}\exp\left(\frac{A^{\pi_k}(s,a)}{\tau}\right) \log \pi_\theta(a|s) 
    \right].
\end{equation}

We write $A=A^{\pi_k}$ for brevity. To encourage exploration and prevent premature convergence, we are inspired by the entropy-constrained formulation of SAC~\cite{haarnoja2018sac}. We optimize the policy parameters $\theta$ to maximize the advantage-weighted log-likelihood subject to a minimum entropy constraint $\bar{H}$:
\begin{equation}
\min_\theta \mathcal{J}(\theta) = {\color{NavyBlue}{-\mathbb{E}_{s \sim \mathcal{D}, a\sim \pi_k(\cdot|s)} \left[ \frac{1}{Z(s)}\exp\left(\frac{A(s,a)}{\tau}\right) \log \pi_\theta(a|s) \right]}} \quad \text{s.t.} \quad {\color{ForestGreen}{\mathbb{E}_{s \sim \mathcal{D}}[{H}(\pi_{\theta}(\cdot|s))] \geq \bar{H}}}, 
\label{eq:objective_general}
\end{equation}
where ${H}(\pi_{\theta}(\cdot|s))$ is the policy entropy. We handle this constraint using a Lagrangian formulation:
\begin{equation}
    \mathcal{L}(\theta, \alpha) = {\color{NavyBlue}{-\mathbb{E}_{s \sim \mathcal{D}, a\sim \pi_k(\cdot|s)}\left[ \frac{1}{Z(s)}\exp\left(\frac{A(s,a)}{\tau}\right) \log \pi_\theta(a|s) \right]}} + \alpha {\color{ForestGreen}{\left(\bar{H}- \mathbb{E}_{s \sim \mathcal{D}}[{H}(\pi_{\theta}(\cdot|s))] \right)}},
    \label{eq:lagrangian}
\end{equation}
where $\alpha \geq 0$ is the Lagrange multiplier, dynamically adjusted via dual gradient descent \cite{haarnoja2018sac} to maintain the target entropy $\bar{H}$. The critic estimates the standard discounted reward return without future entropy terms; entropy enters only through the actor constraint. FMER is therefore an entropy-constrained policy optimization method rather than a full maximum-entropy RL algorithm. Appendix~\ref{appendix:why_not_joint} details our rationale for not jointly optimizing policy mirror descent and the entropy constraint.

Computing $\log \pi_\theta(a|s)$ requires solving an ODE, and directly optimizing Eq.~\eqref{eq:lagrangian} by backpropagating gradients through the ODE integration induces training instability~\cite{ding2024diffusion, zhang2026sac}. To address these challenges, we substitute the log-likelihood term with the Weighted Conditional Flow Matching (${\color{NavyBlue}\mathcal{L}_{\text{W-CFM}}}$) loss, which serves as a \textit{simulation-free and tractable} proxy. Together with the entropy regularization term ${\color{ForestGreen}\mathcal{L}_\text{ent}}$ (introduced in Sec.~\ref{sec:entropy_loss}), we define the unified FMER loss:
\begin{equation}
\mathcal{L}_{\text{FMER}}(\theta) = {\color{NavyBlue}\mathcal{L}_{\text{W-CFM}}(\theta)} + \alpha {\color{ForestGreen}\mathcal{L}_{\text{ent}}(\theta)}.
\end{equation}

\begin{figure}[t]
    \centering
    \includegraphics[width=\linewidth]{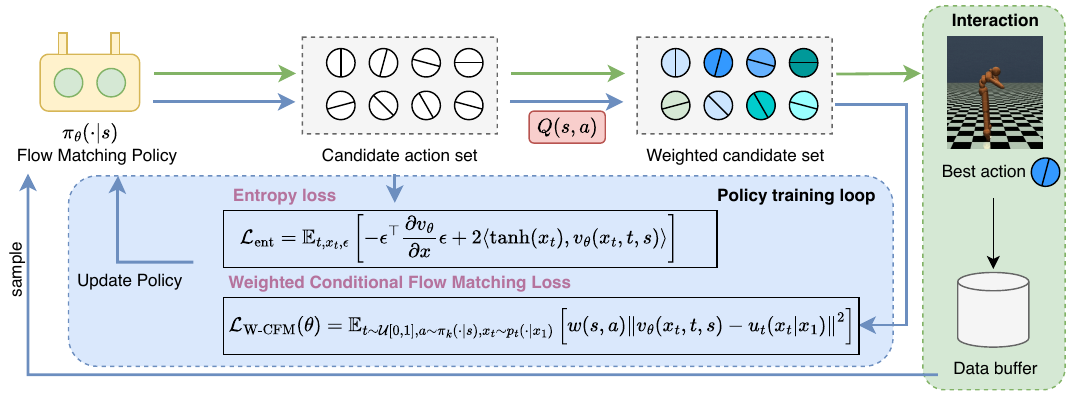}
    \caption{Overview of the FMER framework. Left (Policy Training): 
    The policy generates a set of candidate actions, which are evaluated by the critic (Q-network) to compute advantage-based weights. These weights guide the Weighted Conditional FM loss, while an entropy regularization term ensures exploration.     
    Right (Interaction): During data collection, the agent samples multiple candidates and executes the highest-value action to store in the replay buffer.}
    \label{fig:fmer_diagram}
\end{figure}

The remainder of this section is organized as follows: Sec.~\ref{sec:fm_policy_representation} presents the policy representation, Sec.~\ref{sec:a_weighted_loss} details the ${\color{NavyBlue}\mathcal{L}_{\text{W-CFM}}(\theta)}$ loss, Sec.~\ref{sec:entropy_loss} describes the entropy loss ${\color{ForestGreen}\mathcal{L}_{\text{ent}}(\theta)}$, and Sec.~\ref{sec: training_procedure} presents the complete training procedure. The policy training pipeline is illustrated in Figure~\ref{fig:fmer_diagram}.

\subsection{Flow Matching Policy Representation}
\label{sec:fm_policy_representation}
We parameterize $\pi_\theta(a|s)$ using a latent FM framework. To ensure valid actions in bounded continuous control environments (e.g., $a \in (-1, 1)^d$), we define the flow in an unbounded latent space $x_t \in \mathbb{R}^d$ and obtain actions via $a = \tanh(x_1)$, where $x_1 \sim p_1(\cdot|s)$ and $p_1(\cdot|s)$ represents the \textit{terminal} distribution generated at $t=1$. This guarantees valid actions by design. While squashing actions with a $\tanh$ operator is common in Gaussian policies \cite{haarnoja2018sac, fujimoto2018td3}, computing the entropy of a flow policy requires special care to account for the density distortion induced by this transformation (Section~\ref{sec:entropy_loss}). 

The policy is trained by regressing $v_\theta$ toward the Optimal Transport target velocity $x_1-x_0$ (Eq.~\ref{eq:target_field}-\ref{eq:cfm_loss}), where $x_1 = \text{arctanh}(a)$ and $x_t = (1-t)x_0 + tx_1$:
\begin{equation}
\mathcal{L}_{\pi\text{-CFM}}(\theta) = \mathbb{E}_{t \sim \mathcal{U}[0,1], x_0 \sim p_0, (s, a) \sim \mathcal{D}} \left[ \left\| v_\theta(x_t, t, s) - (\text{arctanh}(a) - x_0) \right\|^2 \right].
\label{eq:general_FM_loss}
\end{equation}

At inference, actions are generated by integrating the ODE via Euler steps \cite{fql_park2025}. The trajectory is discretized into $N$ steps of size $\Delta t = 1/N$ and $x_0 \sim \mathcal{N}(0, \mathbf{I})$:
\begin{equation}
x_{t+\Delta t} = x_{t} + \Delta t \cdot v_\theta(x_t, t, s), \quad t \in \left\{0, \tfrac{1}{N}, \dots, \tfrac{N-1}{N}\right\}.
\label{eq:dis_FM}
\end{equation}
The resulting terminal latent state $x_1$ is then transformed into the executable action via $a = \tanh(x_1)$.

\subsection{Policy Optimization via Weighted-CFM}
\label{sec:a_weighted_loss}
Eq.~\ref{eq:general_FM_loss} can be used for behavior cloning. In online RL, we leverage a learned value function $Q_\psi$ (parameterized by $\psi$) to guide policy updates. As shown in Figure~\ref{fig:fmer_diagram}, at each iteration $k$, we sample actions from the current policy $\pi_k$, evaluate them via $Q_\psi$, and apply advantage-based weights to the Flow Matching loss, steering the vector field toward higher-value regions of the action space.

\begin{theorem}[Weighted Flow Matching Surrogate]
\label{theo:wcfm}
Fix a temperature $\tau>0$, a reference policy $\pi_k$,
and an advantage function $A$. For each state $s$, let
$w(s,a)=\frac{\exp(A(s,a)/\tau)}{Z(s)}$ where
$Z(s):=\mathbb E_{a\sim\pi_k(\cdot\mid s)}
\left[\exp(A(s,a)/\tau)\right]$,
$x_1=\text{arctanh}(a)$, and
$x_t=(1-t)x_0+tx_1$. Consider the Weighted Conditional Flow Matching loss
\begin{equation}
{\color{NavyBlue}\mathcal L_{\mathrm{W\text{-}CFM}}(\theta)
=\mathbb E_{{t \sim \mathcal{U}[0,1], s\sim\mathcal{D}, a \sim \pi_k(\cdot|s), x_0\sim p_0}}
\left[{w(s,a)}\|v_\theta(x_t,t,s)-(x_1-x_0)\|^2\right]}
\label{eq:q_cfm_loss}
\end{equation}
Assume $0<Z(s)<\infty$ for $\mathcal D$-almost every state, and that
the velocity network is expressive enough to represent the target
marginal velocity field.
Under the integrability and regularity conditions, every global minimizer $\theta^*$
induces the ideal policy mirror descent target
\begin{equation}
\pi_{\theta^*}(a\mid s)=\pi^*(a\mid s):=
\frac{\pi_k(a\mid s)\exp(A(s,a)/\tau)}{Z(s)}
\label{eq:pmd_target}
\end{equation}
for $\mathcal D$-almost every $s$ and almost every $a$.
\end{theorem}

\noindent\textbf{Remark.} The weight $w(s, a)$ in Eq.~\eqref{eq:q_cfm_loss} depends only on the terminal action $a=\tanh(x_1)$, see Appendix~\ref{appendix: clarification_w} for why the endpoint weight is the appropriate choice.

The full proof is provided in Appendix~\ref{sec:appendx_A1}. Connecting advantage-weighted policy improvement to vector field regression is non-trivial. Whereas methods such as SAC-Flow \cite{zhang2026sac} attempt to mitigate the gradient instability inherent to backpropagating through the ODE, \cref{theo:wcfm} enables a simulation-free training paradigm that entirely circumvents this issue, highlighting its practical significance.

\begin{wrapfigure}{r}{0.5\textwidth} 
  \centering
  \includegraphics[width=\linewidth]{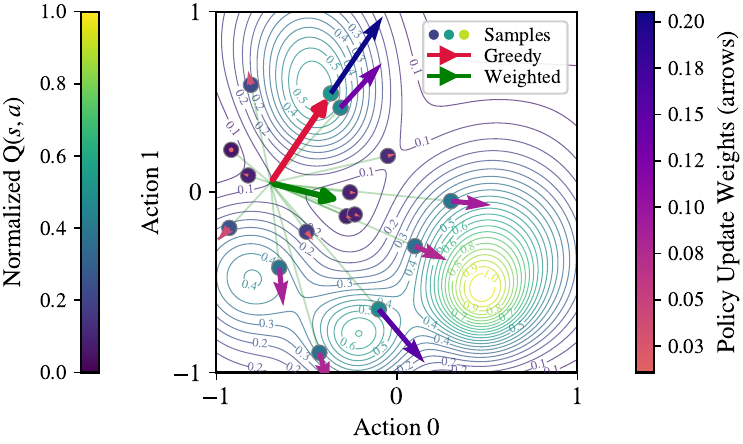}
    \caption{Weighted vs. greedy policy updates. Contours show the ground truth Q-value landscape for a fixed state. From 16 sampled candidate actions, the top-1 (greedy) update moves toward the highest-valued sample (red arrow), leading to a local maximum. In contrast, the weighted update (green) aggregates all candidates via Q-based weights (small arrows) towards the global optimal.
  }
    \label{fig:ov}
\end{wrapfigure}

In practice, we approximate the state-dependent normalizing constant using $M$ candidate actions
$\{a^{(j)}\}_{j=1}^M \sim \pi_k(\cdot|s)$: $\hat{Z}(s)=\frac{1}{M}\sum_{l=1}^M\exp\left(A(s,a^{(l)})/\tau\right).$
Substituting $\hat{Z}(s)$ into the empirical weighted-CFM objective and absorbing the Monte Carlo factor $1/M$
into the weights gives: 
\begin{equation}
    \hat{w}(s, a^{(j)}) =
    \frac{\exp\left(A(s,a^{(j)})/\tau\right)}
    {\sum_{l=1}^M \exp\left(A(s,a^{(l)})/\tau\right)}.
    \label{eq:a-weight}
\end{equation}
Although omitting $Z(s)$ preserves the ideal statewise minimizer under the assumptions of \cref{theo:wcfm},
retaining its estimate normalizes the total candidate weight for each state. Unlike top-1 candidate selection
for policy fitting, this soft weighting retains information from multiple high-quality actions. As illustrated in
Figure~\ref{fig:ov}, top-1 selection discards this information and can be brittle to critic estimation errors such as overestimation.

Since the state-only baseline $V^{\pi_k}(s)$ is constant across candidates, it cancels during normalization.
The weights can therefore be computed directly from $Q_\psi(s,a^{(j)})$, without learning a separate value
network. We discuss connections to advantage-weighted and softmax policy methods in Appendix~\ref{appendix:connection_adv} and connections to policy mirror descent methods in Appendix~\ref{appendix:connection_pmd}.

\paragraph{Adaptive tuning of $\tau$.}
The temperature $\tau$ controls the KL penalty strength in the policy mirror descent objective, governing the strictness of the trust region around $\pi_k$. Large $\tau$ flattens the weights, pulling updates toward the behavior distribution; small $\tau$ sharpens them, allowing more aggressive deviation toward high-advantage actions. We tune $\tau$ adaptively by monitoring the Effective Sample Size (ESS) of the normalized weights $\{\hat{w}_l(s)\}_{l=1}^M$:
\begin{equation}
    \mathrm{ESS}(s) = \frac{1}{\sum_{l=1}^{M} \hat{w}_l(s)^2},
    \label{eq:ess}
\end{equation}
which ranges from $1$ (one weight dominates) to $M$ (uniform weights). We select a target $\mathrm{ESS}^* \in (1, M)$. Similar to the update of $\alpha$, we adjust $\tau$ at each step to maintain the target $\mathrm{ESS}^*$: decreasing $\tau$ when $\mathrm{ESS} > \mathrm{ESS}^*$ (overly uniform weights, constraint too strict) and increasing it when $\mathrm{ESS} < \mathrm{ESS}^*$ (few actions dominate, constraint too loose).

\subsection{Entropy-Regularized Framework} 
\label{sec:entropy_loss}
The stochastic reverse process of diffusion-based policies renders their entropy intractable. In Section~\ref{sec:fm_policy_representation}, we apply a nonlinear transformation $a = \tanh(x_1)$ to map from the unbounded FM latent space into the bounded action space. The quantity of interest is the entropy of \textit{action} instead of the latent $x_1$. \textit{While prior work either ignores or approximates the probability distortion introduced by this change of variables, our work closes this gap with a closed-form formulation.} 

\begin{theorem}[Entropy of Latent Flow, restated from \cite{chen2018neural, DBLP:conf/iclr/LipmanCBNL23} for the OT path]
\label{theo:entropy_lf}
Consider a latent vector field $v_\theta(x, t, s)$ that transports a base distribution $p_0$ to the target latent distribution $p_1$ at $t=1$. For the unbounded latent space $\mathbb{R}^d$, the entropy $H^x(p_1(\cdot|s))$ can be computed as:
\begin{equation}
H^x_1(p_1) = H^x_0 + \int_0^1 \mathbb{E}_{x_t \sim p_t(\cdot|s)} \left[ \nabla_x \cdot v_\theta(x_t, t, s) \right] dt,
\label{eq:H_1_form}
\end{equation}
where $H^x_0 = \frac{d}{2}(1 + \ln(2\pi))$ is the entropy of the base Gaussian noise $p_0 = \mathcal{N}(0, \mathbf{I})$, $d$ is the latent dimension, and $\nabla_x \cdot v_\theta = \text{Tr}(\nabla_x v_\theta)$ is the divergence of the vector field.
\end{theorem}

The proof is provided in Appendix~\ref{sec:appendx_A2}. This theorem provides a tractable mechanism to control policy exploration directly through the vector field divergence. To ensure simulation-free optimization consistent with ${\color{NavyBlue}\mathcal{L}_{\text{W-CFM}}}$, we evaluate divergence along the Optimal Transport path: $x_t = (1 - t)x_0 + tx_1$. By sampling $t \sim \mathcal{U}[0,1]$, the entropy is unbiasedly estimated by:
\begin{equation}
H_1^x \approx H_0^x + \mathbb{E}_{t \sim \mathcal{U}[0,1], x_t \sim p_t(\cdot|s)} \left[ \nabla \cdot v_\theta(x_t, t, s) \right].
\label{eq:H_1^x}
\end{equation}

\begin{proposition}[Divergence under Tanh Transformation]
\label{prop:divergence_tanh}
Consider the coordinate transformation $a = \tanh(x)$ mapping any latent state $x \in \mathbb{R}^d$ to the bounded action space $(-1, 1)^d$. Let $v_\theta(x, t, s)$ be the vector field in the latent space. The corresponding induced vector field in the action space, $v^a(a, t, s)$, satisfies the following divergence relationship at any point $x$:
\begin{equation}
\nabla_a \cdot v^a(a, t, s) = \nabla_x \cdot v_\theta(x, t, s) - \sum_{i=1}^d 2 \tanh(x_i) v_{\theta,i}(x, t, s),
\label{eq:divergence_tanh}
\end{equation}
where $v_{\theta,i}$ is the $i$-th component of the latent vector field.
\end{proposition}

The proof of \cref{prop:divergence_tanh} can be found in Appendix~\ref{sec:appendx_A3}. Eq.~\eqref{eq:divergence_tanh} allows us to optimize the action-space entropy $\pi(a|s)$ by adjusting the latent flow. To maximize entropy (minimize negative entropy), we define the entropy regularization loss $\mathcal{L}_{\text{ent}}$ as:
\begin{equation}
{\color{ForestGreen}\mathcal{L}_{\text{ent}} 
=-\mathbb{E}_{t, x_t}\left[\nabla_x \cdot v_\theta(x_t, t, s) - 2\langle \tanh(x_t), v_\theta(x_t, t, s)\rangle \right].}
\label{eq:loss_ent}
\end{equation}

Appendix~\ref{appendix:connection_maxent} details the relationship between our FMER and existing MaxEnt methods. In practice, computing the divergence $\nabla_x \cdot v_\theta(x) = \sum_{i=1}^d \frac{\partial v_{\theta,i}}{\partial x_i}$ in Eq.~\eqref{eq:loss_ent} scales as $\mathcal{O}(d^2)$. To maintain the computational efficiency necessary for high-dimensional benchmarks, we use an unbiased Hutchinson trace estimator \citep{DBLP:conf/iclr/GrathwohlCBSD19, hutchinson1989stochastic}. This reduces the per-step complexity to $\mathcal{O}(d)$, ensuring training remains computationally feasible (see Appendix~\ref{appendix:Hutchinson} for details). Hutchinson estimation is only used for actor training and not for action generation, training and inference costs are reported in Appendix~\ref{appendix:training_time}.

\noindent\textbf{Remark.} Since $H_0^a$ depends only on the fixed base distribution $p_0$ via a constant Jacobian correction to $H_0^x$, it is constant w.r.t. $\theta$ and can be omitted from the optimization objective (see Appendix~\ref{appendix:H_0aremark} for the full derivation).

\begin{algorithm}[!t]
\caption{Flow Matching Policy Optimization with Mirror Descent and Entropy Constraints}
\label{alg:fmer}
\begin{algorithmic}[1]
   \STATE {\bfseries Input:} Parameters $\theta, \psi, \alpha,\tau$; Target entropy $\bar{H}$; Replay buffer $\mathcal{D}$.
   \FOR{each epoch}
       \STATE \textbf{// Interaction}
       \STATE Observe $s$; generate $M$ candidates $\{a^{(j)}\}_{j=1}^M$ via ODE (Eq.~\ref{eq:dis_FM}); select $a$ with best $Q$ value.
       \STATE Execute $a$, observe $r, s'$, store $(s, a, r, s')$ in $\mathcal{D}$.
       \STATE \textbf{// Training}
       \FOR{$K$ gradient steps}
           \STATE Sample batch $B \sim \mathcal{D}$.
           \STATE \textit{Critic:} Select best of $M$ actions; compute targets $y$ (Eq.~\ref{eq:target_TD}); update $\psi$ (Eq.~\ref{eq:loss_q}).
           \STATE \textit{Policy:} Sample $M$ actions from $\pi_\theta(\cdot|s)$, compute weights $w$ (Eq.~\eqref{eq:a-weight}), sample flow time $t$; update $\theta$ minimizing $\mathcal{L}_{\text{W-CFM}} + \alpha \mathcal{L}_{\text{ent}}$ (Eqs.~\ref{eq:q_cfm_loss}, \ref{eq:loss_ent}).
           \STATE Update $\alpha$; update $\tau$; update target networks.
       \ENDFOR
   \ENDFOR
\end{algorithmic}
\end{algorithm}

\subsection{Training Procedure}
\label{sec: training_procedure}
The complete procedure is summarized in Algorithm~\ref{alg:fmer}. For the value network update, the critic parameters $\psi$ are updated to minimize the Bellman Mean Squared Error (MSE). Given a transition tuple $(s_i, a_i, r_i, s_{i+1})$ sampled from the replay buffer $\mathcal{D}$, the loss is defined as:
\begin{equation}
    \mathcal{L}_Q(\psi) = \mathbb{E}_{\mathcal{D}} \left[ \left( y_i - Q_{\psi}(s_i, a_i) \right)^2 \right],
    \label{eq:loss_q}
\end{equation}
where $y_i$ is the regression target constructed using the target networks $\bar{Q}_{\bar{\psi}_1}, \bar{Q}_{\bar{\psi}_2}$. To mitigate value overestimation, we employ the Clipped Double Q-learning mechanism~\cite{fujimoto2018td3}:
\begin{equation}
    y_i = r_i + \gamma \min_{j=1,2} \bar{Q}_{\bar{\psi}_j}(s_{i+1}, a_{i+1}), 
    \label{eq:target_TD}
\end{equation}
where $a_{i+1}^* = \arg \max_{a^{(m)}} \min_{j=1,2} \bar{Q}_{\bar{\psi}_j}(s_{i+1}, a^{(m)})$ and $\{a^{(m)}\}_{m=1}^M \sim \pi_\theta(\cdot|s_{i+1})$. As detailed in Section~\ref{sec:a_weighted_loss}, we optimize the weighted FM loss using $M$ sampled actions. During data collection, following \cite{ding2024diffusion, kang2023efficient}, we execute the highest-valued of $M$ candidates to interact with the environment. We store the resulting transition in the replay buffer to ensure the replay buffer accumulates high-reward transitions. Similarly, we compute the target for the critic update (Eq.~\ref{eq:target_TD}) using the highest-valued candidate to reduce Bellman backup variance.

\section{Experiments}
\label{sec:experiments_}
In this section, we first visualize the multimodal expressiveness of FMER on a simple multi-goal environment, then evaluate performance on long-horizon sparse-reward multi-task FrankaKitchen tasks and dense-reward MuJoCo benchmarks. Finally, we present ablation studies to quantify the individual contribution of each component.

\begin{wrapfigure}{r}{0.45\textwidth} 
    \centering
    \includegraphics[width=\linewidth]{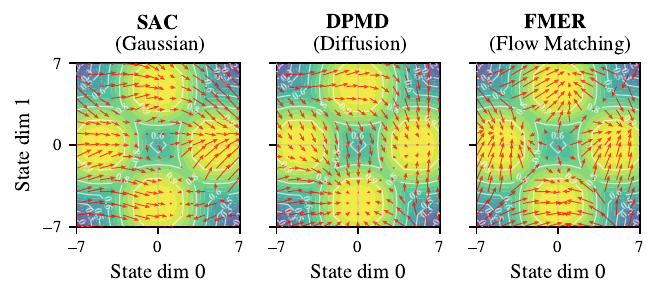}
    \caption{State-conditioned policy visualization on the 2D multi-goal task. Background colors display the ground-truth state-value landscape, with bright areas marking optimal goals. Red arrows show the selected actions for SAC~\cite{haarnoja2018sac}, DPMD~\cite{ma2025efficientonlinereinforcementlearning}, and FMER, illustrating each method's navigation strategy.}
    \label{fig:policy_vis}
\end{wrapfigure}
\subsection{Visualizing Policies for \textit{Multi-Goal}}
To compare the multimodal capacity of different stochastic policy families, we use the standard 2D multi-goal environment commonly adopted for interpretability studies~\cite{haarnoja2017reinforcement}. We compare SAC (Gaussian policy), DPMD (diffusion policy), and FMER (our flow matching policy). Figure~\ref{fig:policy_vis} shows state-conditioned action selection across the environment after convergence. As indicated by the red arrows, SAC and DPMD consistently favor a subset of optimal regions (right and bottom), even when alternative peaks are equally optimal. In contrast, FMER adapts to the local geometry and points toward the nearest of the four equal-valued maxima. This indicates a diverse multimodal behavior facilitated by the tractable entropy control of our ODE-based formulation. Further visualization of the action generation process is provided in Appendix~\ref{appendix:multi_goal}.

\subsection{Benchmarks on FrankaKitchen and Mujoco}
\label{sec:benchmark_franka_mujoco}
\begin{figure}[!t]
    \centering
    \begin{subfigure}[b]{0.15\textwidth}
        \centering
        \includegraphics[width=\textwidth]{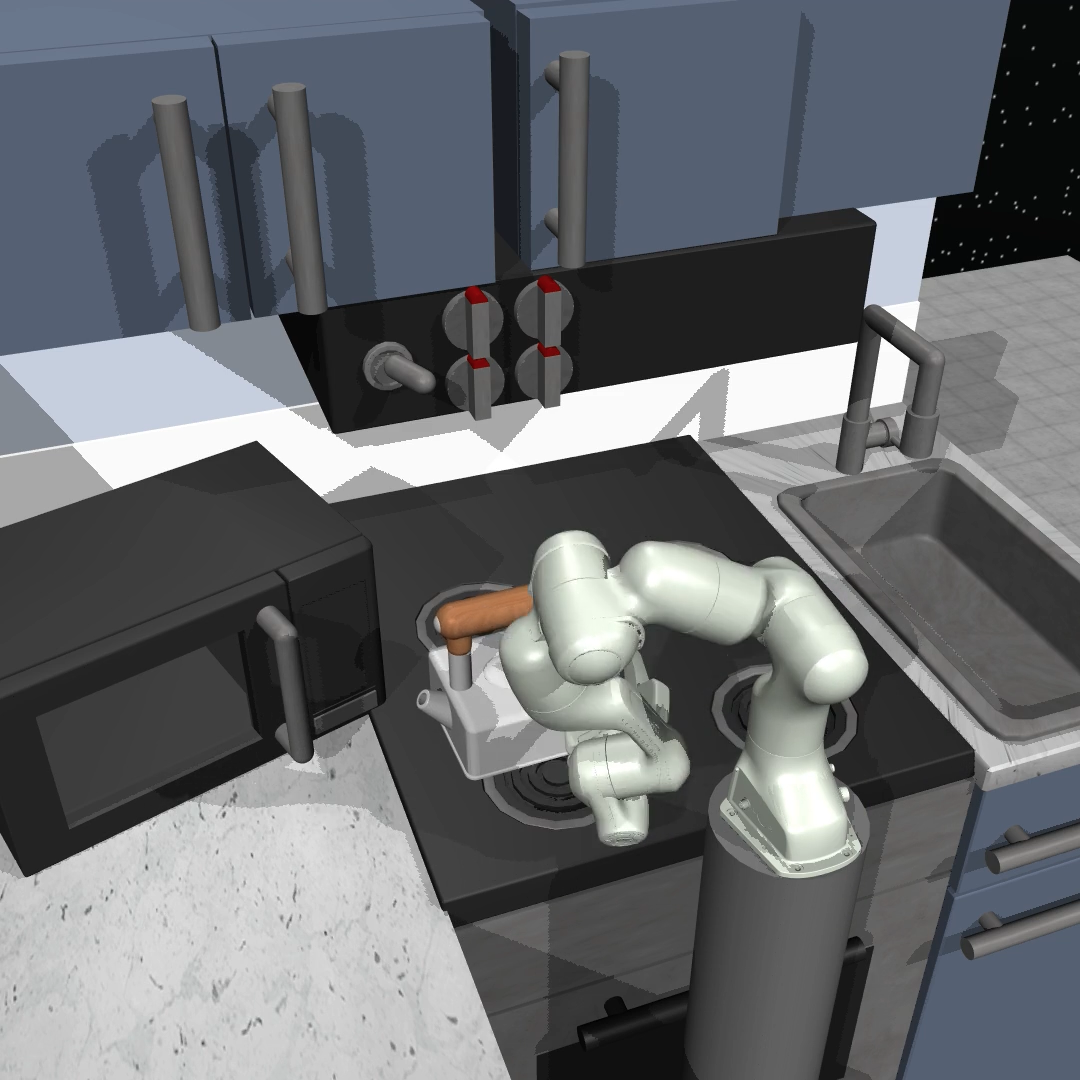}
        \caption{Kettle}
    \end{subfigure}\hfill
    \begin{subfigure}[b]{0.15\textwidth}
        \centering
        \includegraphics[width=\textwidth]{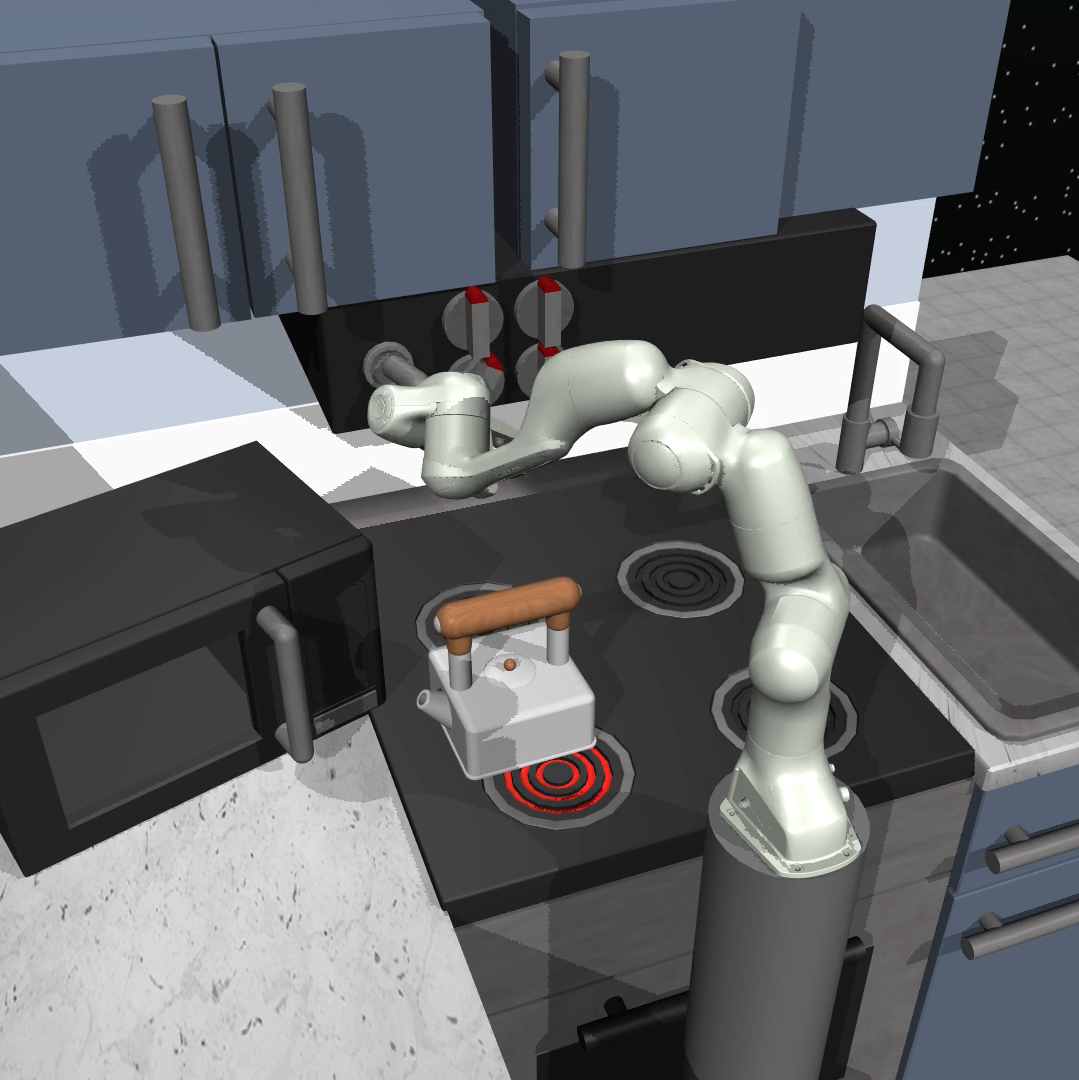}
        \caption{Bot. burner}
    \end{subfigure}\hfill
    \begin{subfigure}[b]{0.15\textwidth}
        \centering
        \includegraphics[width=\textwidth]{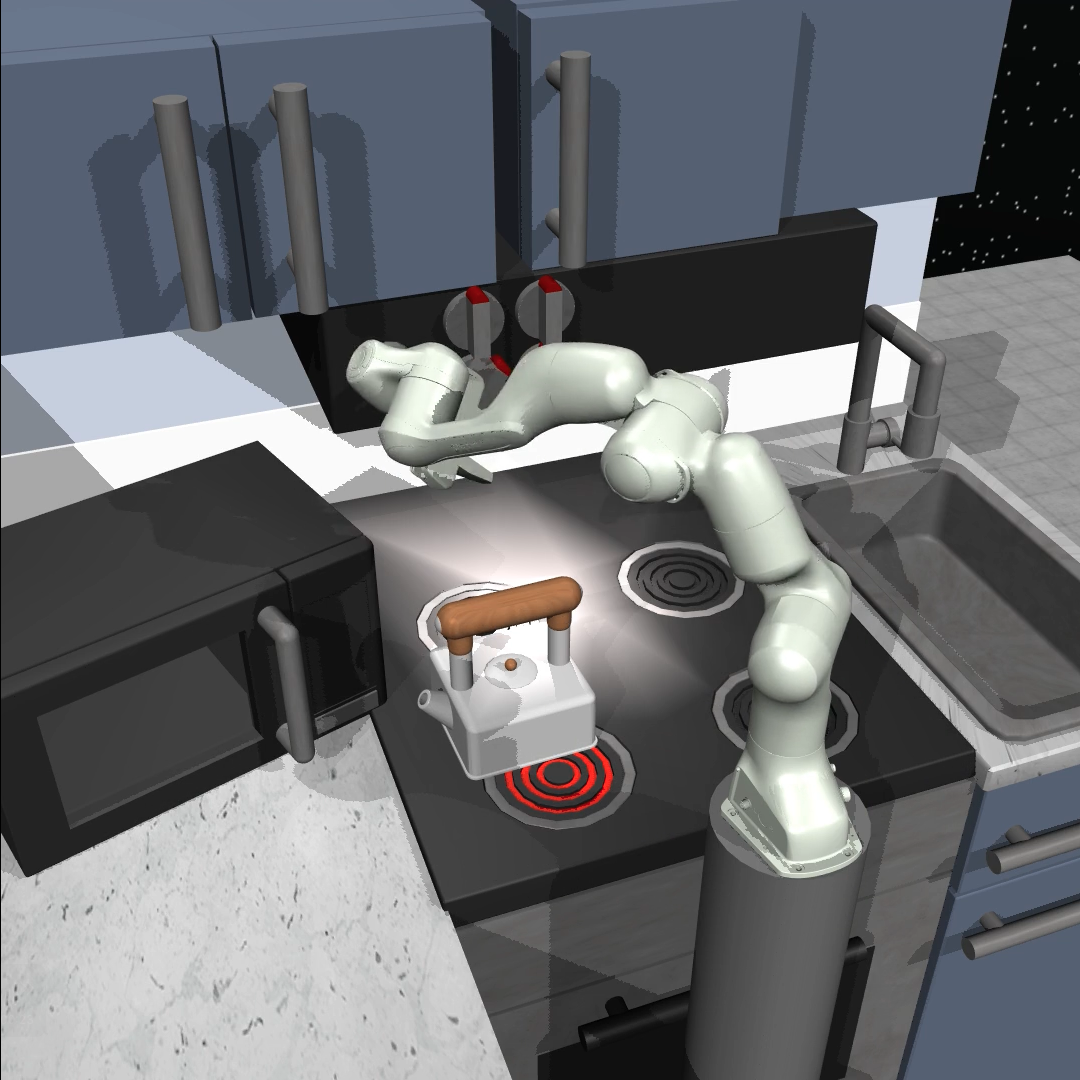}
        \caption{Light switch}
    \end{subfigure}\hfill
    \begin{subfigure}[b]{0.15\textwidth}
        \centering
        \includegraphics[width=\textwidth]{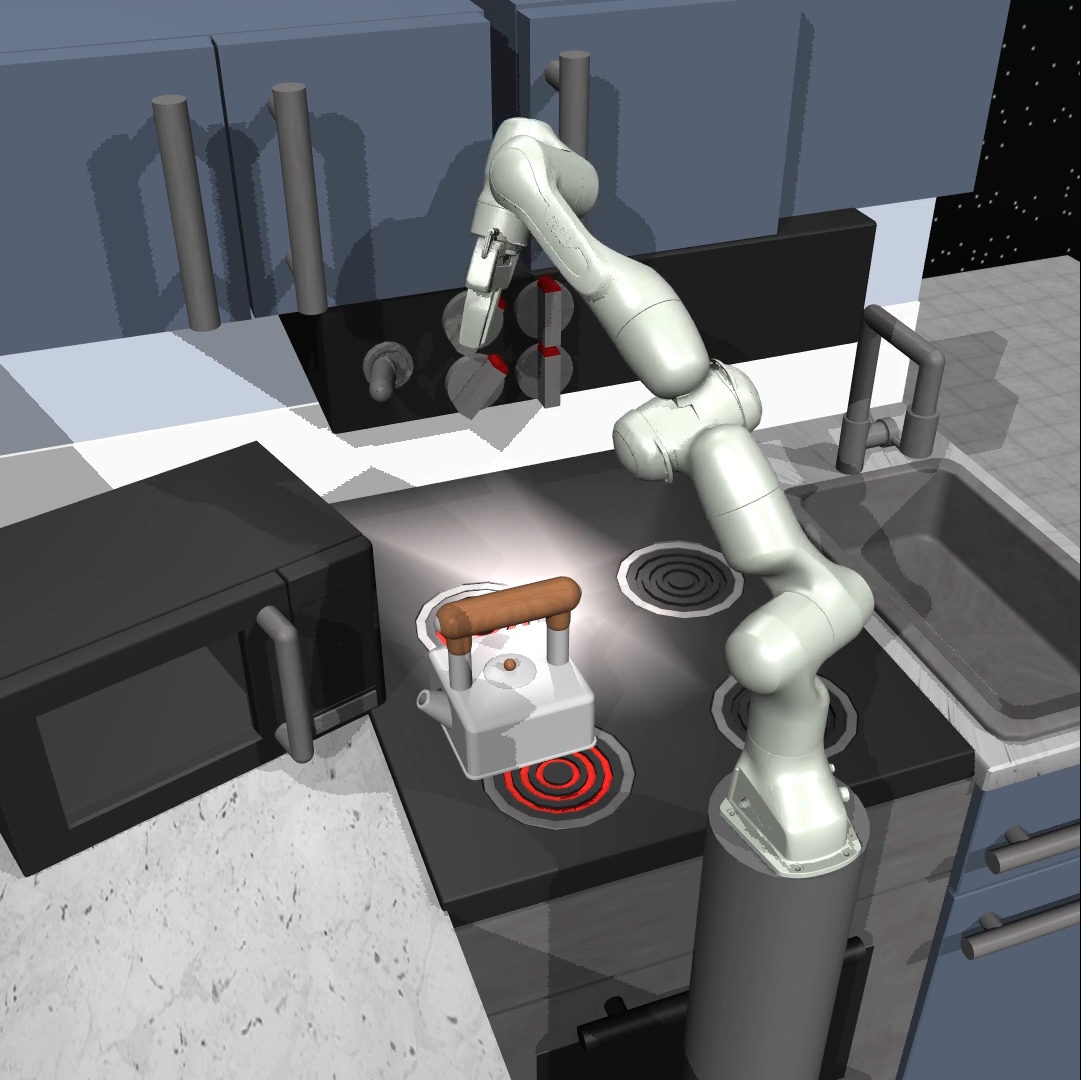}
        \caption{Top burner}
    \end{subfigure}\hfill
    \begin{subfigure}[b]{0.15\textwidth}
        \centering
        \includegraphics[width=\textwidth]{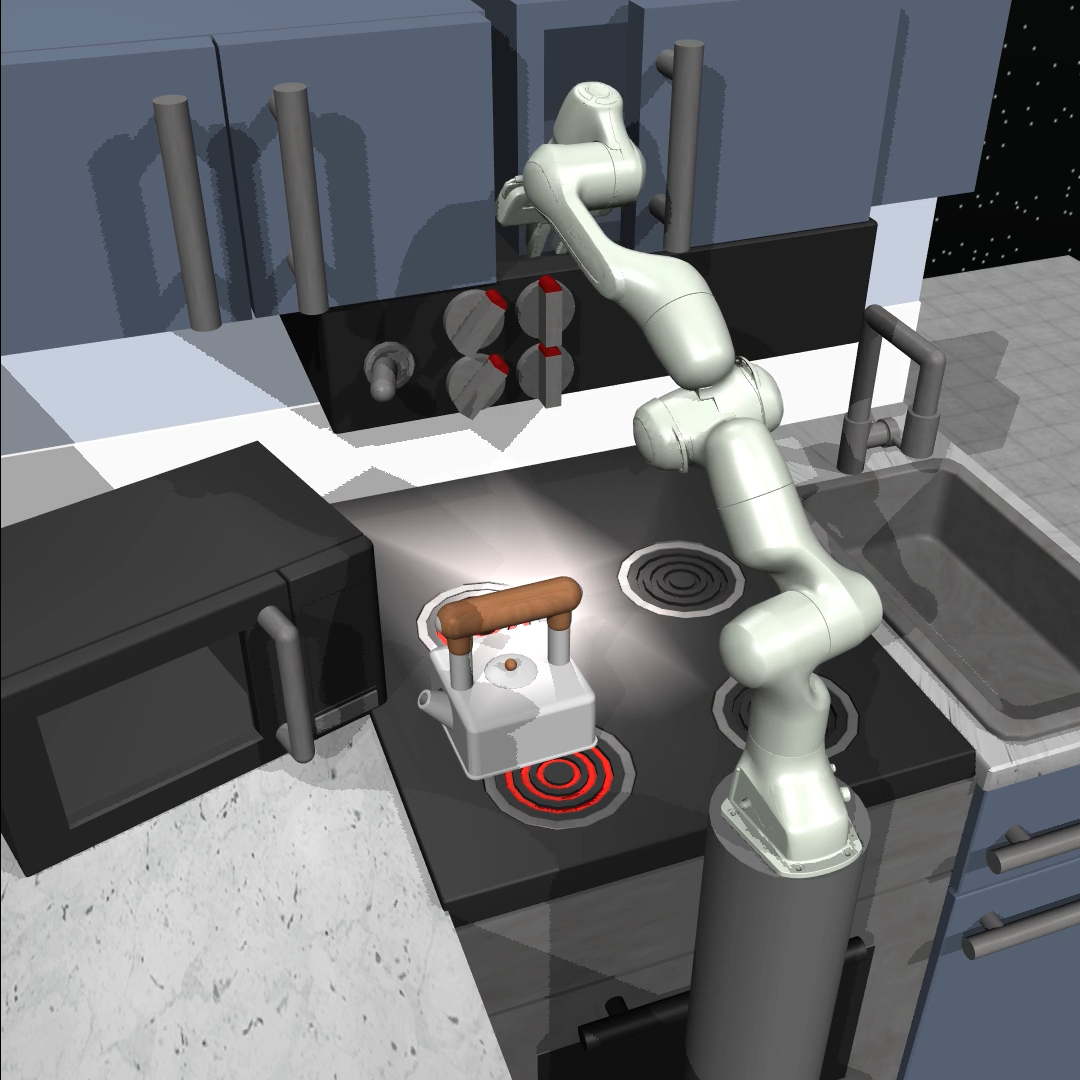}
        \caption{Slide cabinet}
    \end{subfigure}

    \vspace{0.5em} 

    \begin{subfigure}[b]{0.15\textwidth}
        \centering
        \includegraphics[width=\textwidth]{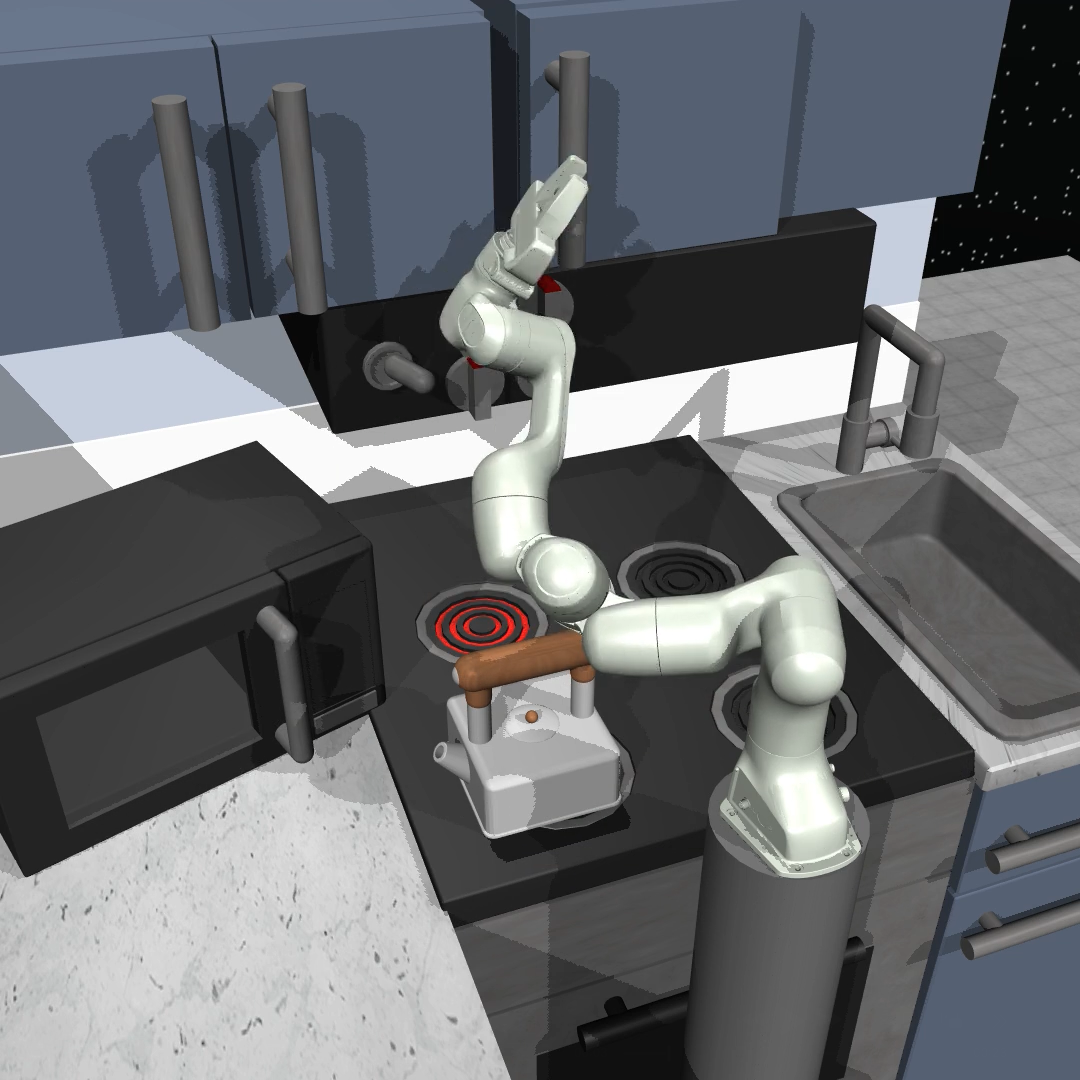}
        \caption{Top burner.}
    \end{subfigure}\hspace{0.0125\textwidth}%
    \begin{subfigure}[b]{0.15\textwidth}
        \centering
        \includegraphics[width=\textwidth]{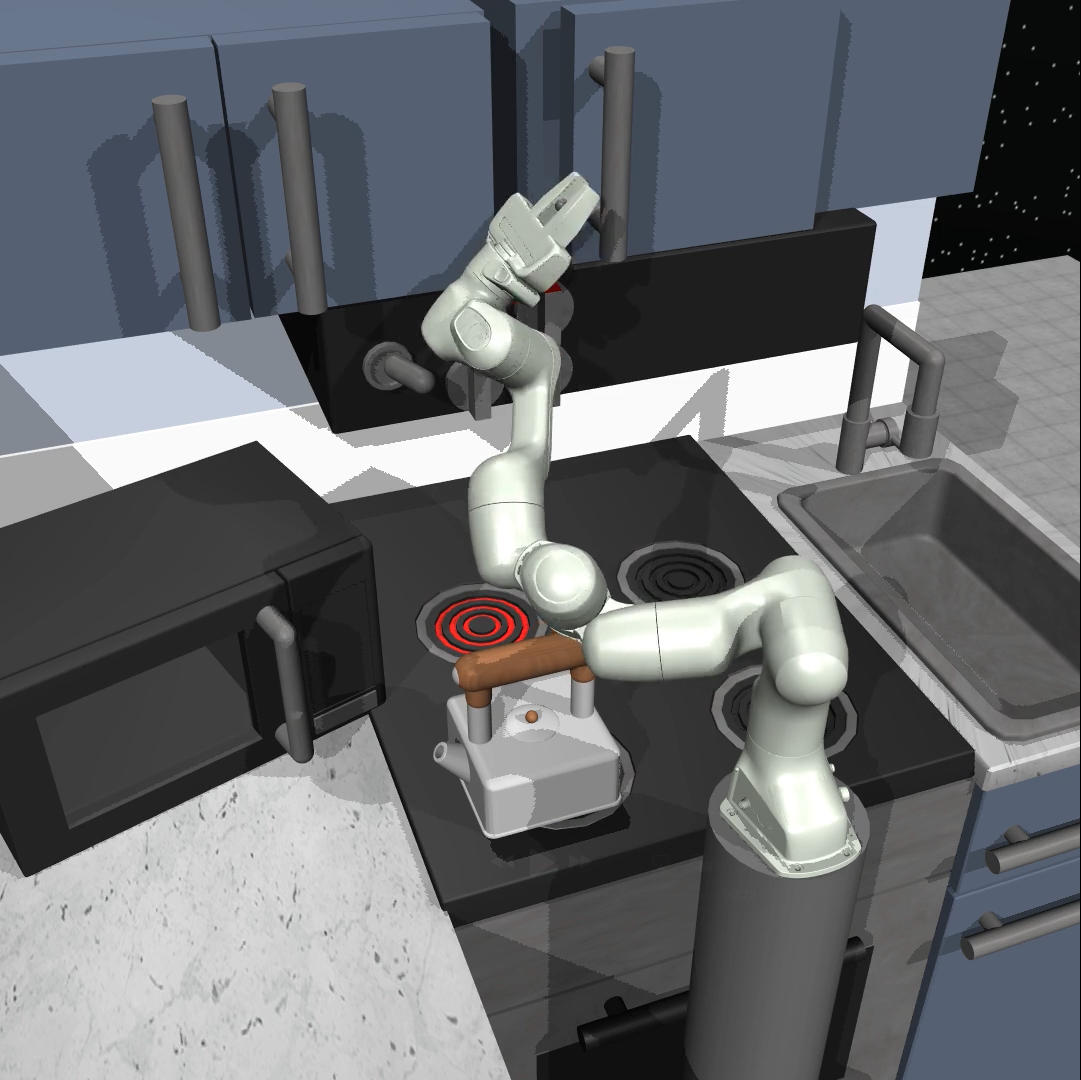}
        \caption{Slide cabinet}
    \end{subfigure}\hspace{0.0125\textwidth}%
    \begin{subfigure}[b]{0.15\textwidth}
        \centering
        \includegraphics[width=\textwidth]{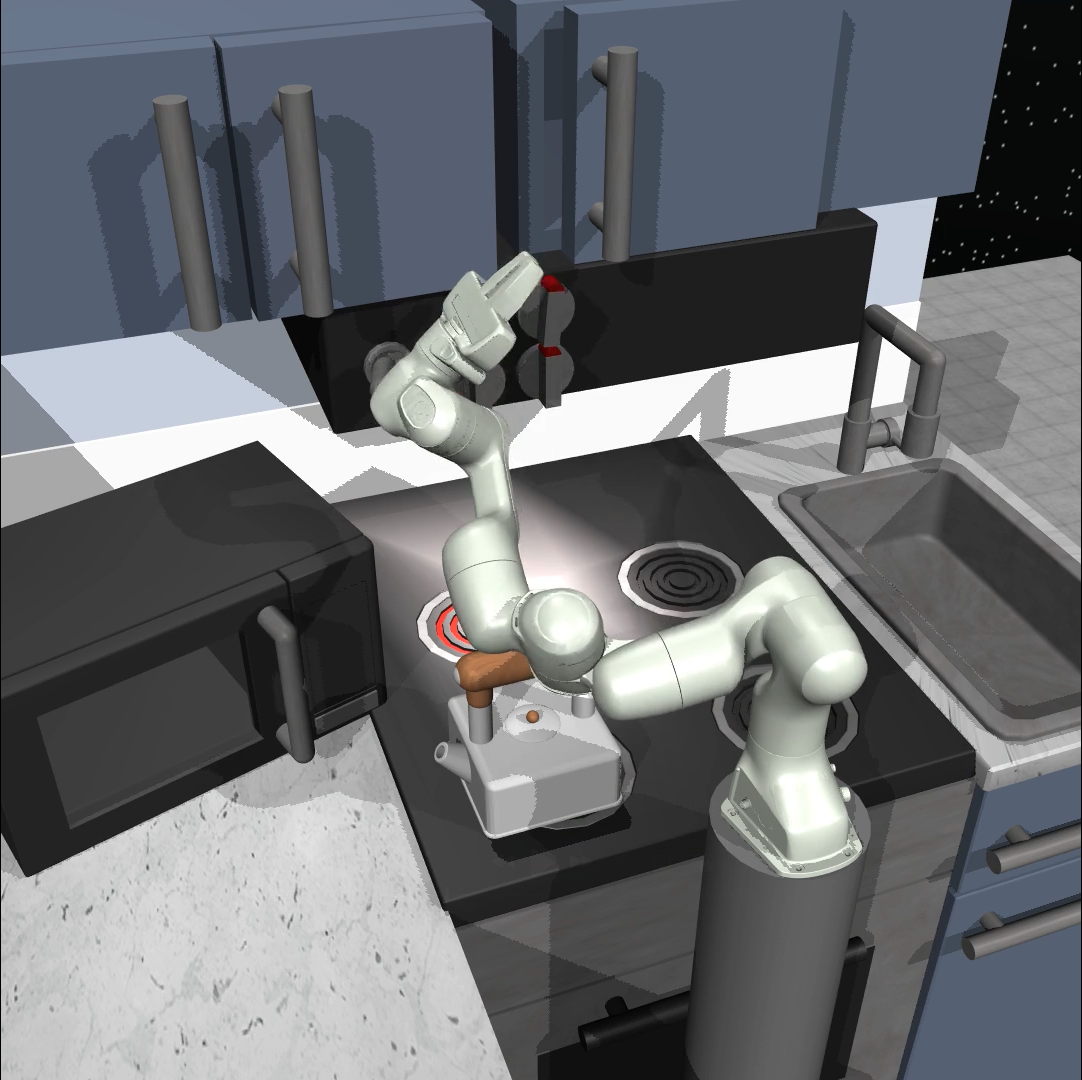}
        \caption{Light switch}
    \end{subfigure}

    \caption{Agents working in the default FrankaKitchen environment with a budget of 280 steps. Top row (a-e): FMER is able to finish 5 tasks. Bottom row (f-h): The best baseline model (DIPO) is able to finish 3 tasks. In this visualization, we use the best seed for both FMER and DIPO.}
    \label{fig:main_span_figure}
\end{figure}

We evaluate FMER against eight baselines across FrankaKitchen~\cite{pmlr-v100-gupta20a} and MuJoCo \cite{towers2024gymnasium}, spanning standard RL (PPO~\cite{schulman2017ppo}, SAC~\cite{haarnoja2018sac}, TD3~\cite{fujimoto2018td3}), diffusion-based (DIPO~\cite{dipo}, QVPO~\cite{ding2024diffusion}, DPMD~\cite{ma2025efficientonlinereinforcementlearning}), and flow-based (SAC-Flow~\cite{zhang2026sac}, FLAME \cite{li2026boostingmaximumentropyreinforcement}) methods. Table~\ref{tab:concept_comparison} provides a conceptual comparison; full hyperparameters are in Appendix~\ref{appendix:baseline}--\ref{appendix:fmer_hp}. All methods are trained for 1M environment steps over 5 random seeds. Off-policy methods include a 10K-step random warm-up and all policies are evaluated every 10K steps over 10 independent episodes. Results report mean maximum return over the final 10\% of training (steps 900K--1M) following \cite{dacer}; training time comparisons are in Appendix~\ref{appendix:training_time}.

\begin{table}[!t]
\small
\centering
\caption{Normalized task completion rate on FrankaKitchen environment and performance on MuJoCo-v5 environment. Values are presented as Mean (Std) across 5 random seeds. For each seed, the policy is evaluated over 10 independent episodes.}
\label{tab:kitchen_mujoco}
\resizebox{\linewidth}{!}{%
\begin{tabular}{l|cccc|cccc}
\toprule
&\multicolumn{4}{c|}{\textbf{FrankaKitchen (Task Completion Rate)}} &\multicolumn{4}{c}{\textbf{MuJoCo-v5}} \\
\textbf{Method} & \textbf{N=1} & \textbf{N=2} & \textbf{N=4} & \textbf{N=7} & \textbf{Hopper} & \textbf{HalfCheetah} & \textbf{Walker} & \textbf{Humanoid} \\
\midrule
\textbf{PPO} & 2.00 (4.47) & 0.00 (0.00) & 1.00 (2.24) & 8.57 (7.82) & 1198.02 (360.64) & 5982.74 (810.60) & 4598.21 (222.17) & 811.94 (78.20)\\
\textbf{SAC} & 40.00 (54.77) & 32.00 (24.90) & 2.50 (3.54) & 12.29 (4.24) & 3318.64 (132.80) & 10295.33 (1170.43) & 4441.51 (536.85) & \textbf{5326.73 (58.93)} \\
\textbf{TD3} & 0.00 (0.00) & 11.00 (24.60) & 15.00 (17.32) & 5.71 (7.82) & \textbf{3662.14 (61.58)} & 9496.59 (901.14) & 4887.65 (630.96) & 318.33 (208.02) \\
\textbf{DIPO}  & 0.00 (0.00) & 33.00 (30.33) & 14.50 (21.39) & 23.14 (18.36) & 3408.50 (297.31) & 9516.47 (1417.20) & 5028.66 (217.36) & 5087.46 (100.04)\\
\textbf{DPMD} & 60.00 (54.77) & \textbf{50.00 (0.00)} & 25.00 (17.68) & 20.57 (8.67) & 3397.57 (278.06) & 10706.06 (817.59) & 4901.24 (351.59) & 5084.60 (27.90)\\
\textbf{QVPO}  & 0.00 (0.00) & 0.00 (0.00) & 0.00 (0.00) & 18.86 (9.44) & 3422.73 (413.41) & 10309.60 (1398.37) & 4846.39 (397.71) & 5064.21 (48.82)\\
\textbf{SAC-Flow} & 0.00 (0.00) & 0.00 (0.00) & 0.00 (0.00) & 0.57 (1.28) & 2930.95 (510.33) & 9933.18 (254.50) & 5233.78 (728.87) & 4857.14 (2763.20)\\
\textbf{FLAME} & 8.00 (17.89) & 10.00 (22.36) & 6.50 (14.53) & 18.29 (7.79) & 2593.95 (432.33) & 10900.69 (1478.33) & 4407.20 (143.72) & 5207.44 (345.98)\\
\midrule
\textbf{FMER} &\textbf{80.00 (44.72)} & 48.00 (4.47) & \textbf{46.50 (3.79)} & \textbf{40.29 (21.03)} & 2913.45 (642.67) & \textbf{12332.38 (903.67)} & \textbf{5282.89 (437.57)} & 5286.12 (65.35)\\
\bottomrule
\end{tabular}%
}
\end{table}
\noindent\textbf{FrankaKitchen.} The FrankaKitchen benchmark~\cite{pmlr-v100-gupta20a} features a 9-DoF Franka robot with sparse rewards across interactable kitchen objects, evaluated on four task configurations $N \in \{1,2,4,7\}$ (setup details in Appendix~\ref{appendix:frankakitchen_setup}). The environment is sparse: the agent has 280 steps per episode and receives a binary reward of 1 only upon completing a sub-task. It also exhibits a multimodal value distribution, as multiple distinct interaction sequences can satisfy the same goal. As shown in Table~\ref{tab:kitchen_mujoco} and Figure~\ref{fig:main_span_figure}, FMER outperforms all baselines across nearly all configurations with increasingly superior results at higher $N$.

Gaussian policies (PPO, SAC, TD3) struggle with the non-Gaussian optimal behavior distribution; SAC degrades from $N=2$ to $N=4$ as action multimodality grows beyond what a unimodal policy can capture. Among generative baselines, DPMD's KL trust region locks onto early successes but prevents escape from local optima at higher $N$; QVPO's uniform-distribution entropy objective and SAC-Flow's biased entropy surrogate both limit effective exploration in sparse settings. FMER's tractable entropy and flow-matching backbone enable principled exploration and multimodal target tracking throughout, yielding consistently superior performance. Full results and analysis, including an additional comparison with the on-policy method FPO~\cite{fmpo}, are provided in Appendix~\ref{appendix:franka_perf}.

\noindent\textbf{MuJoCo-v5.} Following prior diffusion- and FM-based work \cite{zhang2026sac, ding2024diffusion, ma2025efficientonlinereinforcementlearning}, we include MuJoCo as a regression test for whether FMER's multimodal machinery incurs unnecessary costs in simpler settings. 
As shown in Table~\ref{tab:kitchen_mujoco}, FMER attains the highest returns on HalfCheetah and Walker2d, and outperforms all other generative methods on Humanoid (trailing only SAC). However, FMER underperforms off-policy baselines on Hopper. We attribute this to the entropy constraint sustaining exploration longer than necessary in this relatively simple unimodal environment. Training curves and detailed analysis are provided in Appendix~\ref{appendix:mujoco_curves}.

\subsection{Ablation study}
\noindent\textbf{Ablation on Best-of-$M$ Action Selection.} As described in Section~\ref{sec: training_procedure}, FMER selects the highest-valued action among $M$ candidates for both environment interaction and critic-target computation. 
\begin{wraptable}{r}{0.5\textwidth}
    \centering
    \caption{Ablation on best-of-$M$ action selection on FrankaKitchen. Task completion rates (\%) are presented as Mean (Std).}
    \label{tab:no_argmax}
    \resizebox{\linewidth}{!}{%
        \begin{tabular}{lcccc}
            \toprule
            Method & $N=1$ & $N=2$ & $N=4$ & $N=7$ \\
            \midrule
            DPMD, default & 60.00 (54.77) & \textbf{50.00 (0.00)} & 25.00 (17.68) & 20.57 (8.67) \\
            FMER, default & 80.00 (44.72) & 48.00 (4.47) & \textbf{46.50 (3.79)} & \textbf{40.29 (21.03)} \\
            \midrule
            DPMD, no argmax & 0.00 (0.00) & 0.00 (0.00) & 5.00 (11.18) & 0.86 (1.28) \\
            FMER, no argmax & \textbf{100.00 (0.00)} & 50.00 (35.36) & 26.50 (2.24) & 14.29 (0.00) \\
            \bottomrule
        \end{tabular}%
    }
\end{wraptable}
To examine whether this selection mechanism explains FMER's advantage, we compare it with DPMD, the strongest PMD-based baseline on FrankaKitchen. Both methods use best-of-$M$ selection by default, with $M=32$ for DPMD and $M=8$ for FMER. We evaluate both methods under their default settings and a matched no-argmax ablation, in which a single action is sampled for environment interaction and critic-target computation, removing both argmax operations. The candidate sets used for actor training remain unchanged. 

We remark that, although $N=1$ offers fewer reward opportunities, it is also the simplest task. As shown in Table~\ref{tab:no_argmax}, candidate argmax does not fully explain FMER’s advantage over DPMD. Under the matched no-argmax setting, FMER continues to learn across all task configurations, whereas DPMD barely succeeds. Moreover, removing candidate argmax does not reduce FMER's mean performance on $N=1$ or $N=2$, but substantially reduces it on $N=4$ and $N=7$. This comparison complements the entropy ablation: it rules out candidate argmax as the sole explanation for the performance gap, while acknowledging its substantial contribution on the harder long-horizon settings, likely because it helps collect rewarding transitions and produces stronger critic targets.

\begin{wrapfigure}{r}{0.45\textwidth} 
    \centering
    \includegraphics[width=0.7\linewidth]{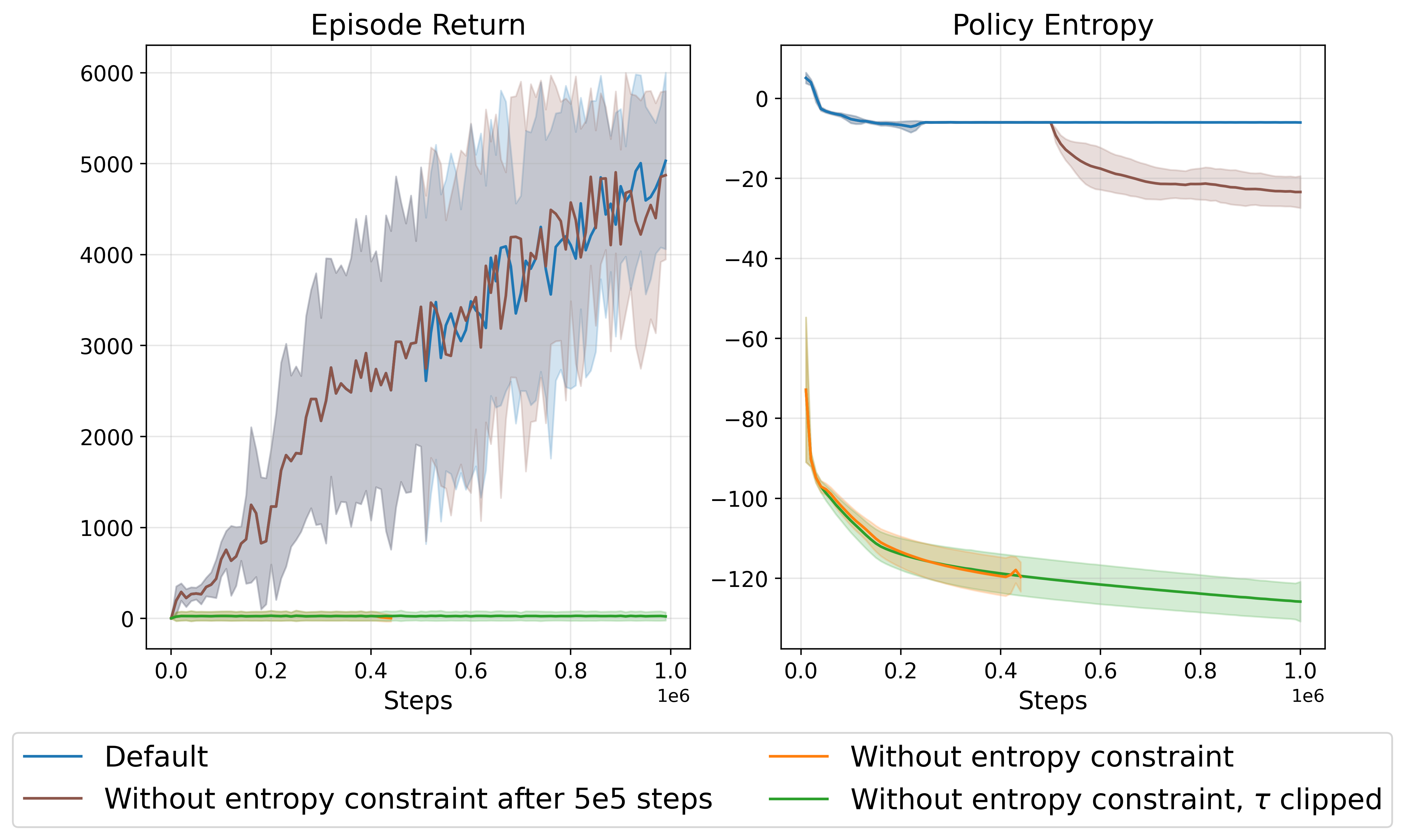}
    \caption{Training curves (left) and entropy evolution (right) when removing the entropy constraint after $m$ steps on Walker2d-v5.}
    \label{fig:entropy}
\end{wrapfigure}
\noindent\textbf{Entropy constraint.}  In Figure~\ref{fig:entropy}, we demonstrate the necessity of entropy regularization by deactivating $\mathcal{L}_{\text{ent}}$ at different $m$ training steps. When deactivated from the start ($m=0$), the policy collapses to near-deterministic behavior: $\mathrm{ESS}$ keeps being above $\mathrm{ESS}^*$, driving $\tau \to 0$. Although clipping $\tau$ from below prevents numerical instability, it cannot avert the underlying policy collapse. In contrast, 
deactivating the constraint late in training ($m=5 \times 10^5$) has limited impact, as sufficient exploration has already been established. In both cases, entropy drops immediately upon deactivation, confirming the efficiency of entropy control with $\alpha$. Appendix~\ref{appendix:tanh} validates the theoretical and practical necessity of the $\tanh$ correction (\cref{prop:divergence_tanh}), and Appendix~\ref{appendix:abl_Hutchinson} evaluates the computational trade-offs of exact trace calculation. 

\begin{wrapfigure}{rt}{0.45\textwidth} 
    \centering
    \includegraphics[width=0.7\linewidth]{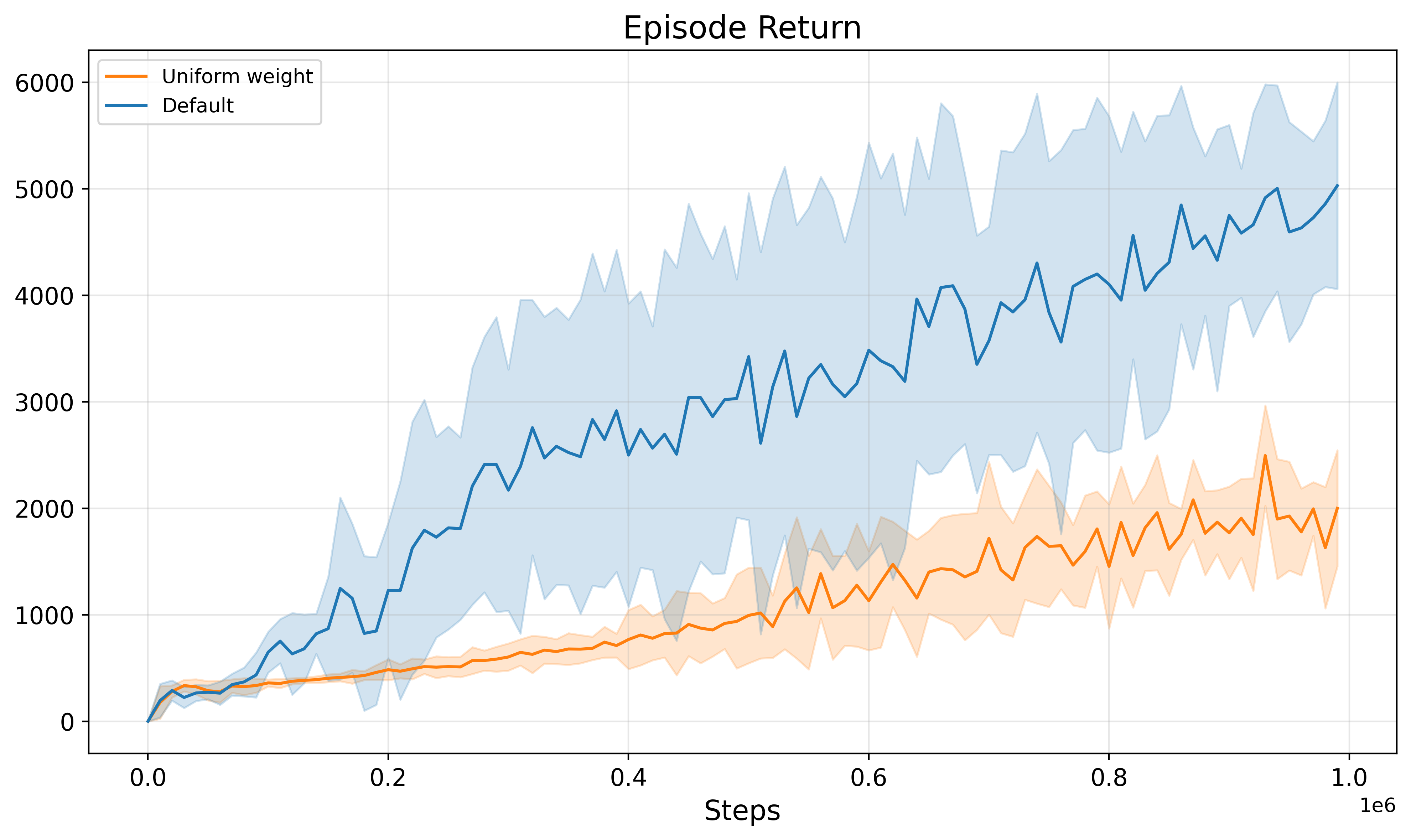}
    \caption{Training curves comparing advantage-based weight and uniform weight on Walker2d-v5.}
    \label{fig:uniform_weight}
\end{wrapfigure}
\noindent\textbf{Advantage-weighting ablations.} Figure~\ref{fig:uniform_weight} ablates the mirror descent weighting by replacing the soft advantage weights in Eq.~\eqref{eq:q_cfm_loss} with uniform weights, reducing the loss to standard behavior cloning over all sampled actions regardless of quality. This significantly harms the performance: assigning equal weight to high-quality and low-quality actions pulls the policy toward mediocre behavior, confirming that exponential advantage weighting is a necessary mechanism for directed policy improvement rather than a soft heuristic. Appendix~\ref{appendix:adaptive_tau} shows that $\mathrm{ESS}$-based adaptive $\tau$ avoids environment-specific hyperparameter tuning. We provide further ablation studies on candidate size in Appendix~\ref{appendix:abl_candidate}, and on ODE steps and solvers in Appendix~\ref{appendix:ode_steps_solver}. 

We repeat both the entropy and advantage-weighting ablations on sparse-reward FrankaKitchen ($N=7$), with results reported in Appendix~\ref{appendix:abl_franka_ent_adv}.

\section{Conclusion}
We presented FMER, an FM policy optimization framework for online RL that addresses two fundamental challenges: policy expressiveness and the exploration-exploitation tradeoff. We show that the policy mirror descent update can be realized via advantage-weighted conditional flow matching loss, enabling simulation-free training without backpropagating through the ODE. We further derive a closed-form tractable entropy expression which explicitly corrects for the density distortion induced by the $\tanh$ transformation from the unbounded latent space to the bounded action space. Empirical results validate each contribution: the 2D multi-goal environment confirms the multimodal expressiveness of the flow matching policy, FrankaKitchen and MuJoCo benchmarks demonstrate that FMER achieves superior or competitive performance across sparse- and dense-reward settings, and ablation studies isolate the necessity of both mirror descent weighting and closed-form entropy control. Limitations and future work are discussed in Appendix~\ref{sec:appendix_limitation}.

\clearpage

\section*{Acknowledgment}
This work was jointly supported by the EMERALDS project under the Horizon Europe Research and Innovation Programme (Grant Agreement No. 101093051) and the Transport \& Mobility Institute at Delft University of Technology. The authors acknowledge the use of computational resources of the DelftBlue supercomputer, provided by Delft High Performance Computing Centre. This work also used the Dutch national e-infrastructure with the support of the SURF Cooperative using grant no. EINF-16178.

\small
\bibliographystyle{plainnat}
\bibliography{bibliography}


\newpage
\appendix
\addtocontents{toc}{\protect\setcounter{tocdepth}{2}}
\tableofcontents

\section{Proofs}
\label{sec:appendix_proofs}

\subsection{Proof of \cref{theo:wcfm}}
\label{sec:appendx_A1}

\textbf{\cref{theo:wcfm} (Weighted Flow Matching Surrogate).}
Fix a temperature $\tau>0$, a reference policy $\pi_k$,
and an advantage function $A$. For each state $s$, let
$w(s,a)=\frac{\exp(A(s,a)/\tau)}{Z(s)}$ where
$Z(s):=\mathbb E_{a\sim\pi_k(\cdot\mid s)}\left[\exp(A(s,a)/\tau)\right]$,
$x_1=\text{arctanh}(a)$, and
$x_t=(1-t)x_0+tx_1$. Consider the Weighted Conditional Flow Matching loss
\begin{equation}
\mathcal L_{\mathrm{W\text{-}CFM}}(\theta)
=\mathbb E_{{t \sim \mathcal{U}[0,1], s\sim\mathcal{D}, a \sim \pi_k(\cdot|s), x_0\sim p_0}}
\left[{w(s,a)}\|v_\theta(x_t,t,s)-(x_1-x_0)\|^2\right].
\label{eq:loss_wcfm}
\end{equation}
Assume $0<Z(s)<\infty$ for $\mathcal D$-almost every state, and that
the velocity network is expressive enough to represent the target
marginal velocity field.
Under the integrability and regularity conditions, every global minimizer $\theta^*$
induces the ideal policy mirror descent target
\begin{equation}
\pi_{\theta^*}(a\mid s)=\pi^*(a\mid s):=
\frac{\pi_k(a\mid s)\exp(A(s,a)/\tau)}{Z(s)}
\end{equation}
for $\mathcal D$-almost every $s$ and almost every $a$.

\begin{proof}
Eq.~\eqref{eq:ideal_pmd_policy} in
Appendix~\ref{appendix:pmd_proof} identifies $\pi^*$ as
the ideal Policy Mirror Descent (PMD) optimizer:
\begin{equation}
\pi^*(a\mid s)=\frac{1}{Z(s)} \pi_k(a\mid s)\,\exp\left(\frac{A(s,a)}{\tau}\right)=w(s,a)\pi_k(a\mid s).
\label{eq:pmd_reweighting}
\end{equation}
Let $T=\text{arctanh}$ and define
$q_1^*(\cdot\mid s)=T_\#\pi^*(\cdot\mid s)$ as the
distribution of $x_1=T(a)$ for $a\sim\pi^*(\cdot\mid s)$.

We assume that actions lie in $(-1,1)^d$, that
$\mathbb E_{s\sim\mathcal D}\,\mathbb E_{x_1\sim q_1^*(\cdot\mid s)}\|x_1\|^2<\infty$,
and that each $v_\theta$ is regular enough for its ODE flow and continuity equation
to have unique solutions on $[0,1]$.

\noindent\textbf{Reweighting the endpoint distribution.}
The advantage weights can be absorbed into the endpoint
distribution. Substituting Eq.~\eqref{eq:pmd_reweighting} into
Eq.~\eqref{eq:loss_wcfm}, we obtain
\begin{equation}
\begin{aligned}
\mathcal L_{\mathrm{W\text{-}CFM}}(\theta)
&=
\mathbb E_{s\sim\mathcal D}
\left[
\int \pi_k(a\mid s)w(s,a)\,
\mathbb E_{t,x_0}
\left[
\|v_\theta(x_t,t,s)-(T(a)-x_0)\|^2
\right] da
\right]
\\
&=
\mathbb E_{s\sim\mathcal D}
\left[
\int \pi^*(a\mid s)\,
\mathbb E_{t,x_0}
\left[
\|v_\theta(x_t,t,s)-(T(a)-x_0)\|^2
\right] da
\right]
\\
&=
\mathbb E_{s\sim\mathcal D}
\left[
\mathbb E_{t,x_0,\,x_1\sim q_1^*(\cdot\mid s)}
\left[
\|v_\theta(x_t,t,s)-(x_1-x_0)\|^2
\right]
\right].
\end{aligned}
\label{eq:wcfm_reweighted}
\end{equation}
The last equality follows because sampling $a\sim\pi^*(\cdot\mid s)$ and setting $x_1=T(a)$ is equivalent to sampling $x_1\sim q_1^*(\cdot\mid s)$. Throughout, $t\sim\mathcal U[0,1]$, $x_0\sim p_0$, and the endpoint are sampled independently, conditional on $s$. Thus, weighting samples from $\pi_k$ by $w/Z$ yields the
same population objective as unweighted Conditional Flow Matching (CFM) with the PMD target $\pi^*$, without requiring direct samples from $\pi^*$.

Define the inner conditional flow matching loss by
\begin{equation}
\mathcal L_{\mathrm{CFM},s}(v):=\mathbb E_{t,x_0,x_1}\left[\|v(x_t,t,s)-(x_1-x_0)\|^2\right].
\end{equation}
Then Eq.~\eqref{eq:wcfm_reweighted} becomes
\begin{equation}
\mathcal L_{\mathrm{W\text{-}CFM}}(\theta)=\mathbb E_{s\sim\mathcal D}\left[\mathcal L_{\mathrm{CFM},s}(v_\theta)\right].
\label{eq:wcfm_mixture}
\end{equation}

\noindent\textbf{Recovering the ideal PMD policy.}
For each fixed state $s$, the inner loss $\mathcal L_{\mathrm{CFM},s}$ in Eq.~\eqref{eq:wcfm_mixture} is the standard CFM objective in \cite{DBLP:conf/iclr/LipmanCBNL23} with endpoint distribution $q_1^*(\cdot\mid s)$.
The assumed second-moment bound on $x_1$ and the finite second moment of the Gaussian prior $p_0$ ensure that
the minimum state-averaged CFM loss is finite. By the conditional-mean decomposition underlying \citep[Theorem~2]{DBLP:conf/iclr/LipmanCBNL23}, the conditional and marginal flow matching objectives differ only by a constant independent of $v$. Consequently, the minimizing field is
\begin{equation}
u_t^*(x,s)=\mathbb E[x_1-x_0\mid x_t=x,t,s],
\end{equation}
whose flow transports $p_0$ to $q_1^*(\cdot\mid s)$
\citep[Theorem~1]{DBLP:conf/iclr/LipmanCBNL23}. Under the network expressiveness assumption, a single network parameter choice attains the minimum of each statewise loss for $\mathcal D$-almost every state. Every global minimizer $\theta^*$ of their average must therefore satisfy $v_{\theta^*}=u_t^*$ almost everywhere along the interpolation paths.

By the assumed uniqueness, the ODE
\begin{equation}
\frac{dX_t}{dt}=v_{\theta^*}(X_t,t,s),
\qquad X_0\sim p_0,
\end{equation}
therefore produces $X_1\sim q_1^*(\cdot\mid s)$.
Transforming the terminal samples back to the action space gives
\begin{equation}
\pi_{\theta^*}(\cdot\mid s)
=\tanh_\#q_1^*(\cdot\mid s)
=\tanh_\#\bigl(T_\#\pi^*(\cdot\mid s)\bigr)
=\pi^*(\cdot\mid s).
\end{equation}
Since $\tanh\circ T$ is the identity, the induced action distribution equals $\pi^*$. Therefore, minimizing the
population weighted CFM loss recovers the ideal PMD policy for $\mathcal D$-almost every state.
\end{proof}

\subsection{Proof of \cref{theo:entropy_lf}}
\label{sec:appendx_A2}
\textbf{\cref{theo:entropy_lf} (Entropy of Latent Flow \cite{chen2018neural, DBLP:conf/iclr/LipmanCBNL23}).} Consider a latent vector field $v_\theta(x, t, s)$ that transports a base distribution $p_0$ to the target latent distribution $p_1$ at $t=1$. For the unbounded latent space $\mathbb{R}^d$, the entropy $H^x(p_1(\cdot|s))$ can be computed as:
\begin{equation}
H^x_1(p_1) = H^x_0 + \int_0^1 \mathbb{E}_{x_t \sim p_t(\cdot|s)} \left[ \nabla_x \cdot v_\theta(x_t, t, s) \right] dt,
\end{equation}
where $H^x_0 = \frac{d}{2}(1 + \ln(2\pi))$ is the entropy of the base Gaussian noise $p_0 = \mathcal{N}(0, \mathbf{I})$ and $d$ is the latent dimension.

\begin{proof}
For simplicity, we use $H_t$ to denote $H_t^x$ in this proof.
For all $x \in \mathbb{R}^d$ and $t \in [0, 1]$, a flow model with vector field $v_t$ and probability path $p_t(x)$ satisfies the continuity equation:
\begin{equation}
    \frac{\partial p_t}{\partial t} = -\nabla \cdot (p_t v_t).
    \label{eq:appendix_continuity}
\end{equation}
The entropy of the distribution $p_t$ is defined as:
\begin{equation}
    H_t = -\int p_t(x) \log p_t(x) \, dx.
\end{equation}
Taking the time derivative:
\begin{equation}
    \frac{d}{dt}H_t = -\int \left(\log p_t(x)\frac{\partial p_t(x)}{\partial t} + \frac{\partial p_t(x)}{\partial t}\right) \, dx.
\end{equation}
Since $\int p_t(x) \, dx = 1$, the second term vanishes:
\begin{equation}
    \frac{d}{dt}H_t = -\int \log p_t(x) \frac{\partial p_t(x)}{\partial t} \, dx.
\end{equation}
Substituting Eq. \eqref{eq:appendix_continuity} into the expression:
\begin{equation}
    \frac{d}{dt}H_t = \int \log p_t(x) \nabla \cdot (p_t v_t) \, dx.
\end{equation}
Assuming the probability mass vanishes at the boundaries, we apply integration by parts. Note that $\nabla \log p_t = \frac{\nabla p_t}{p_t}$:
\begin{equation}
    \frac{d}{dt}H_t = -\int v_t \cdot \nabla p_t \, dx.
\end{equation}
Using the identity $\nabla \cdot (p_t v_t) = p_t \nabla \cdot v_t + v_t \cdot \nabla p_t$ and the integration by parts:
\begin{equation}
    \frac{d}{dt}H_t = \int p_t \nabla \cdot v_t \, dx.
\end{equation}
Therefore, 
\begin{equation}
    \frac{d}{dt}H_t = \mathbb{E}_{x \sim p_t}[\nabla \cdot v_t].
\end{equation}
Integrating over time from $t=0$ to $t=1$:
\begin{equation}
    H_1 = H_0 + \int_0^1 \mathbb{E}_{x \sim p_t}[\nabla \cdot v_t] \, dt.
\end{equation}
Finally, 
\begin{equation}
    H_1 = H_0 + \mathbb{E}_{t \sim \mathcal{U}[0,1], x \sim p_t}[\nabla \cdot v_t]. 
\end{equation}
\end{proof}

\subsection{Proof of \cref{prop:divergence_tanh}}
\label{sec:appendx_A3}
\textbf{\cref{prop:divergence_tanh} (Divergence under Tanh Transformation).}
Consider the coordinate transformation $a = \tanh(x)$ mapping any latent state $x \in \mathbb{R}^d$ to the bounded action space $(-1, 1)^d$. Let $v_\theta(x, t, s)$ be the vector field in the latent space. The corresponding induced vector field in the action space, $v^a(a, t, s)$, satisfies the following divergence relationship at any point $x$:
\begin{equation}
\nabla_a \cdot v^a(a, t, s) = \nabla_x \cdot v_\theta(x, t, s) - \sum_{i=1}^d 2 \tanh(x_i) v_{\theta,i}(x, t, s),
\end{equation}
where $v_{\theta,i}$ is the $i$-th component of the latent vector field.

\begin{proof}
\cref{theo:entropy_lf} assumes an unbounded integration domain where boundary terms vanish. For bounded actions, we consider the domain $a \in (-1, 1)^d$. We employ a change of variables where $a = \tanh(x)$ and the Flow Matching is performed in the unconstrained latent space $x \in \mathbb{R}^d$. The relationship between the velocity fields $v^a(a)$ in the action space and $v_\theta(x)$ in the latent space is given by the chain rule:
\begin{equation}
    v^a_t(a) = \frac{da}{dt} = \frac{da}{dx}\frac{dx}{dt} = (1-a^2) \odot v_\theta(x, t, s).
\end{equation}
Note that here $(1-a^2)$ acts element wise. The divergence in the action space is:
\begin{equation}
    \nabla_a \cdot v^a_t(a) = \nabla_a \cdot \left( (1-a^2) \odot v_\theta(x) \right).
\end{equation}
Using the product rule for divergence $\nabla \cdot (f \mathbf{v}) = \nabla f \cdot \mathbf{v} + f \nabla \cdot \mathbf{v}$:
\begin{equation}
    \nabla_a \cdot v^a_t(a) = \sum_{i=1}^d \left( v_{\theta, i}(x) \frac{\partial (1-a_i^2)}{\partial a_i} + (1-a_i^2) \frac{\partial v_{\theta, i}(x)}{\partial a_i} \right).
    \label{eq:appendix_nabla_a_v_a}
\end{equation}
The first term simplifies to $-2a_i v_{\theta, i}(x)$. For the second term, applying the chain rule yields:
\begin{equation}
    \frac{\partial v_{\theta, i}(x)}{\partial a_i} = \frac{\partial v_{\theta, i}(x)}{\partial x_i} \frac{\partial x_i}{\partial a_i} = \frac{\partial v_{\theta, i}(x)}{\partial x_i} \frac{1}{1-a_i^2}.
    \label{eq:appendix_chain_rule}
\end{equation}
Substituting Eq.~\eqref{eq:appendix_chain_rule} back into Eq.~ \eqref{eq:appendix_nabla_a_v_a}, the term $(1-a_i^2)$ cancels out:
\begin{equation}
    \nabla_a \cdot v^a_t(a) = \sum_{i=1}^d \left( -2a_i v_{\theta, i}(x) + \frac{\partial v_{\theta, i}(x)}{\partial x_i} \right) = \nabla_x \cdot v_\theta(x) - \sum_{i=1}^d 2a_i v_{\theta, i}(x).
\end{equation}
This confirms the relationship stated in \cref{prop:divergence_tanh}.
\end{proof}

\subsection{Derivation of the Policy Mirror Descent closed-form objective}
\label{appendix:pmd_proof}
We follow the same pipeline as in Appendix A, AWR \cite{peng2019advantage}. AWR is an offline RL algorithm, here we focus on online RL. The data buffer $\mathcal{D}$ approximates on-policy samples from $\pi_k$. Our goal is to find a policy $\pi$ that maximizes the expected advantage $A^{\pi_k}(s, a)$ with respect to the current policy $\pi_k$, subject to a trust-region constraint to stabilize learning. Using the expected advantage, we can formulate the following constrained policy search problem:
\begin{equation}
    \arg\max_{\pi} \int_{s} d_{\pi_k}(s) \int_{a} \pi(a|s) A^{\pi_k}(s, a) da\,ds \quad \text{s.t.} \quad   D_{\text{KL}}(\pi(\cdot|s) \| \pi_k(\cdot|s)) \le \epsilon, \int_{a} \pi(a|s) da = 1, \forall s,
    \label{eq:PMD_constraint}
\end{equation}
where $d_{\pi_k}(s)$ as the unnormalized discounted state distribution induced by the policy $\pi_k$ \cite{sutton1998introduction}. 

We relax the hard KL constraint by converting it into a soft constraint with coefficient $\tau$ (acting as a temperature parameter). Following AWR \cite{peng2019advantage}, we adopt a global scalar multiplier $\tau$ rather than a state-dependent $\tau(s)$, which relaxes the pointwise KL constraint to an expected KL constraint over the state distribution $d^{\pi_k}$.

\begin{equation}
\begin{split}
\arg\max_{\pi} \left( \int_{s} d_{\pi_k}(s) \int_{a} \pi(a|s) A^{\pi_k}(s, a) da\,ds \right) + &\tau \left( \epsilon - \int_{s} d_{\pi_k}(s) D_{\text{KL}}(\pi(\cdot|s) \| \pi_k(\cdot|s)) ds \right)\\
\text{s.t.} &\int_{a} \pi(a|s) da = 1, \forall s .
\end{split}
\end{equation}

Next, we form the Lagrangian, with $\tau$ and $\beta = \{\beta_s | \forall s \in \mathcal{S}\}$ corresponding to the Lagrange multipliers:  
\begin{equation}
\begin{split}
\mathcal{L}(\pi, \tau, \beta) = \left( \int_{s} d_{\pi_k}(s) \int_{a} \pi(a|s) A^{\pi_k}(s, a) da\,ds \right) &+ \tau \left( \epsilon - \int_{s} d_{\pi_k}(s) D_{\text{KL}}(\pi(\cdot|s) \| \pi_k(\cdot|s)) ds \right) \\
+\int_{s} \beta_s \left( 1 - \int_{a} \pi(a|s) da \right) ds .
\end{split}
\end{equation}
Differentiating $\mathcal{L}(\pi, \tau, \beta)$ with respect to $\pi(a|s)$ results in:
\begin{equation}
    \frac{\partial \mathcal{L}}{\partial \pi(a|s)} = d_{\pi_k}(s) A^{\pi_k}(s,a) - \tau d_{\pi_k}(s) \log \pi(a|s) + \tau d_{\pi_k}(s) \log \pi_k(a|s) - \tau d_{\pi_k}(s) - \beta_s.
\end{equation}
Setting to zero and solving for $\pi(a|s)$ gives:
\begin{align}
    \log \pi(a|s) = \frac{1}{\tau} A^{\pi_k}(s,a) + \log \pi_k(a|s) - 1 - \frac{1}{d_{\pi_k}(s)} \frac{\beta_s}{\tau} 
    \\
    \pi(a|s) = \pi_k(a|s) \exp\left(\frac{A^{\pi_k}(s,a)}{\tau}\right) \exp\left(- \frac{1}{d_{\pi_k}(s)} \frac{\beta_s}{\tau} - 1\right)
\end{align}
Since $\int_{a} \pi(a|s) da = 1$, the second exponential term is the partition function $\frac{1}{Z(s)}$ that normalizes the conditional action distribution:  
\begin{equation}
    Z(s) = \exp\left(\frac{1}{d_{\pi_k}(s)} \frac{\beta_s}{\tau} + 1\right) = \int_{a'} \pi_k(a'|s) \exp\left(\frac{A^{\pi_k}(s,a')}{\tau}\right) da' 
\end{equation}
The optimal, non-parametric policy is therefore given by:
\begin{equation}
    \pi^*(a|s) = \frac{1}{Z(s)} \pi_k(a|s) \exp\left(\frac{A^{\pi_k}(s,a)}{\tau}\right) 
    \label{eq:ideal_pmd_policy}
\end{equation}
If $\pi$ is represented by a function approximator (e.g., a neural network $\pi_\theta$), the optimal policy $\pi^*$ can be projected onto the manifold of parameterized policies by solving the following supervised regression problem via KL divergence minimization under the data distribution:  
\begin{align}
    &\arg\min_{\theta} \mathbb{E}_{s \sim \mathcal{D}} \left[ D_{\text{KL}}(\pi^*(\cdot|s) \| \pi_\theta(\cdot|s)) \right] \\
    &= \arg\min_{\theta} \mathbb{E}_{s \sim \mathcal{D}} \left[ D_{\text{KL}}\left( \frac{1}{Z(s)} \pi_k(\cdot|s) \exp\left(\frac{A^{\pi_k}(s,\cdot)}{\tau}\right) \Big\| \pi_\theta(\cdot|s) \right) \right] 
    \\&= \arg\max_{\theta} \mathbb{E}_{s \sim \mathcal{D}, a \sim \pi_k} \left[ \frac{1}{Z(s)}\exp\left(\frac{A^{\pi_k}(s,a)}{\tau}\right) \log \pi_\theta(a|s) \right] 
\end{align}
AWR \cite{peng2019advantage} drops $Z(s)$ entirely, which is one source of its instability under heterogeneous Q-scales across states. FMER retains and empirically approximates it via the softmax denominator in Eq. \eqref{eq:a-weight}. Specifically, $\sum_{l=1}^M \exp(A(s,a^{(l)})/\tau)$ serves as a self-normalized Monte Carlo estimate of $M \cdot Z(s)$ over the candidate action set $\{a^{(l)}\}_{l=1}^M \sim \pi_k(\cdot|s)$. This approximation becomes exact as $M \to \infty$. It also preserves the state-dependent normalization that ensures weights sum to one for each state. 
.

\section{Model design}
\subsection{Clarification of \(w(s,a)\) along the latent path}
\label{appendix: clarification_w}
We first clarify that $x_t$ is an auxiliary flow-time state used only for velocity regression; it is not an action executed in the environment, and there is therefore \textit{no independently defined action value $Q(s,x_t)$ (thus we cannot define $w(s,x_t)$)}. Only the terminal output $a=\tanh(x_1)$ is executed and evaluated by the critic. Accordingly, $w(s,a)$ assigns importance to a sampled terminal action and its associated conditional path.

Importantly, using the same weight along one conditional path does not give equal updates to paths leading to good and bad actions. A path ending at a high-advantage action is multiplied by $w(s,a_{\mathrm{high}})$, whereas a path ending at a low-advantage action is multiplied by the smaller $w(s,a_{\mathrm{low}})$, at every sampled flow time.

\begin{itemize}
    \item \textbf{Consistency with the endpoint objective.}
    The use of the same endpoint weight across flow time follows from the endpoint-weighted AWR objective in \cref{theo:wcfm}. As given by Eq.~\eqref{eq:pmd_reweighting}, the target policy is derived by $w(s,a)$ \textit{depending only on the terminal action}. Replacing $w(s,a)$ with a separate time- or latent-dependent weight $w_t(s,x_t,a)$ would define a different path-dependent objective and would no longer correspond to the same endpoint-weighted policy update. 

    \item \textbf{Overlapping conditional paths.}
    Different flow matching paths can pass through the same latent region and prescribe different target velocities. Standard Conditional Flow Matching resolves this through the marginal vector field. For a fixed $(x,t,s)$, consider
    \begin{equation}
        \ell(v)
        =
        \mathbb{E}\!\left[
            w(s,a)\left\|v(x,t,s)-(x_1-x_0)\right\|^2
            \,\middle|\, x_t=x,t,s
        \right].
    \end{equation}
    Differentiating with respect to $v$ and setting the derivative to zero yields the optimal solution
    \begin{equation}
        v_w^*(x,t,s)
        =
        \frac{
            \mathbb{E}\!\left[
                w(s,a)(x_1-x_0)
                \,\middle|\, x_t=x,t,s
            \right]
        }{
            \mathbb{E}\!\left[
                w(s,a)
                \,\middle|\, x_t=x,t,s
            \right]
        }.
    \end{equation}
    Therefore, when paths toward high- and low-value actions overlap, the learned vector field does not branch. Instead, it aggregates their target velocities, with higher-advantage endpoints contributing more strongly.
\end{itemize}

\subsection{Why not a joint optimization?}
\label{appendix:why_not_joint}
We acknowledge that our current Lagrangian in Eq.~\eqref{eq:lagrangian} decouples the constraints. However, we specifically avoid joint optimization to maintain the benefits of simulation-free training. Following the policy mirror descent in Eq.~\eqref{eq:PMD_constraint}, if we were to include the entropy constraint directly into the primary maximization problem, the objective would become:
\begin{equation}
\begin{split}
    &\arg\max_{\pi} \int_{s} d_{\pi_k}(s) \int_{a} \pi(a|s) A^{\pi_k}(s, a) da\,ds \\
    &\text{s.t.} \quad \mathbb{E}_{s \sim \mathcal{D}}[{H}(\pi(\cdot|s))]\geq \bar{H}, \quad D_{\text{KL}}(\pi(\cdot|s) \| \pi_k(\cdot|s)) \le \epsilon, \quad \int_{a} \pi(a|s) da = 1, \forall s.
\end{split}
\end{equation}

For this joint maximization, we form the Lagrangian with $\tau, \alpha$, and $\beta = \{\beta_s | \forall s \in \mathcal{S}\}$ corresponding to the respective Lagrange multipliers:
\begin{equation}
\begin{split}
    \mathcal{L}(\pi, \tau, \alpha, \beta) = &\left( \int_{s} d_{\pi_k}(s) \int_{a} \pi(a|s) A^{\pi_k}(s, a) da\,ds \right) + \tau \left( \epsilon - \int_{s} d_{\pi_k}(s) D_{\text{KL}}(\pi(\cdot|s) \| \pi_k(\cdot|s)) ds \right) \\ &+\alpha \left( \mathbb{E}_{s \sim \mathcal{D}}[{H}(\pi(\cdot|s))] - \bar{H} \right) + \int_{s} \beta_s \left( 1 - \int_{a} \pi(a|s) da \right) ds. 
\end{split}
\end{equation}

Note that ${H}(\pi) = -\int \pi(a|s) \log \pi(a|s) da$. Similar to Appendix \ref{appendix:pmd_proof}, taking the derivative with respect to $\pi(a|s)$ yields the optimal policy $\pi^*$:
\begin{equation}
    \pi^*(a|s) \propto \pi_k(a|s)^{\frac{\tau}{\alpha+\tau}} \exp\left(\frac{A^{\pi_k}(s,a)}{\alpha+\tau}\right) 
\end{equation}
Without the entropy constraint ($\alpha=0$), the base policy $\pi_k(a|s)$ has an exponent of 1 and can be absorbed into importance sampling. However, with $\alpha>0$, the fractional exponent $\frac{\tau}{\alpha+\tau}$ requires the explicit evaluation of $\pi_k(a|s)$. In our framework, this would require simulating the full backward ODE at every single training step, fundamentally destroying the simulation-free advantage of Conditional Flow Matching (CFM).

\subsection{Connections to Advantage-Weighted and Softmax Policy Methods}
\label{appendix:connection_adv}

Several prior works employ advantage-weighted or softmax-style objectives for diffusion- and flow-based policies. We clarify the key connections and distinctions below.

\begin{itemize}
\item \textbf{AWR~\citep{peng2019advantage}.} As discussed in Section~\ref{sec:optimization_objective}, our advantage-weighted policy objective shares the same trust-region motivation as AWR. The key distinction lies in optimization: directly applying AWR's update to a flow matching policy would require backpropagating through the full ODE integration chain, which is both computationally expensive and numerically unstable. We instead optimize a simulation-free surrogate by sampling $t \sim \mathcal{U}(0,1)$ and regressing the vector field without rolling out the ODE, with \cref{theo:wcfm} providing the formal justification.
\item \textbf{QVPO~\cite{ding2024diffusion}.} QVPO maximizes
$\mathbb{E}_{s,a \sim \pi_k(\cdot|s)}\bigl[Q(s,a)\log\pi_\theta(a|s)\bigr]$,
which requires $Q(s,a) \geq 0$ and it is an assumption that can be violated in general. QVPO addresses this via engineering fixes (\texttt{qcut}, \texttt{qadv}) whose guarantees do not transfer back to the original objective. By contrast, our weights arise directly from policy mirror descent (Section~\ref{sec:a_weighted_loss}): using advantage values $A(s,a)$ rather than raw $Q$-values renders the surrogate relationship in \cref{theo:wcfm} unconditional, requiring no assumption on reward sign or value scale.

\item \textbf{IDQL~\cite{hansen2023idql}.} IDQL applies value-function reweighting only at \emph{inference time} via greedy argmax selection, while the diffusion model itself is trained as a pure behavior cloner without any value-function weighting. In our work, we integrate softmax-normalized advantage weights directly into the \emph{training} objective at every gradient step, coupling policy improvement with representation learning throughout training.

\item \textbf{SDAC~\citep{ma2025efficientonlinereinforcementlearning}.} Among the two methods proposed in~\cite{ma2025efficientonlinereinforcementlearning}, SDAC is the one that most resembles a softmax policy $\pi^*(a|s)\propto\exp(Q(s,a)/\lambda)$, whose normalizing constant $Z = \int \exp(Q(s,a)/\lambda)\,da$ is intractable. During training, SDAC samples $a_t$ from an arbitrary proposal (e.g., uniform), draws noisy estimates $\tilde{a}_0 \sim \mathcal{N}(c_t^1 a_t,\, c_t^2)$ using diffusion coefficients $c_t^1, c_t^2$, and applies exponential weighting to these \emph{noisy} samples; normalization serves only as a numerical stability fix. In our work, softmax weights are applied directly to clean candidate actions sampled from the current policy $\pi_k$, and are grounded in the advantage-weighted log-likelihood objective of \cref{theo:wcfm}. This provides a consistent, state-wise learning signal that is invariant to the absolute scale of $Q$-values across states.
\end{itemize}

\subsection{Connections to MaxEnt methods}
\label{appendix:connection_maxent}
Recent flow-based methods approach entropy calculation in various ways. We note that FLAME~\cite{li2026boostingmaximumentropyreinforcement} is a concurrent and independent work. Below, we contrast our method with these existing techniques:
\begin{itemize}
    \item \textbf{ReinFlow \cite{zhang2026reinflow}}. Designed for fine-tuning pre-trained policies, ReinFlow controls entropy by injecting noise into existing probability paths. However, in from-scratch online training, this noise injection provides no meaningful exploration signal and leads to policy collapse.
    \item \textbf{SAC-Flow \cite{zhang2026sac}}. Whereas SAC-Flow addresses gradient instability during backpropagation through time (BPTT), our approach eliminates the need for BPTT by utilizing simulation-free training. SAC-Flow estimates entropy via a noise-augmented joint path density. This approach overestimates the true policy entropy, causing the optimizer to perceive the policy as more stochastic than it actually is, thereby diminishing the effective exploration pressure. While this bias may provide a stabilizing effect in dense-reward settings, it fundamentally restricts exploration in sparse-reward environments, as confirmed by our experiments in Section~\ref{sec:benchmark_franka_mujoco}: SAC-Flow collapses entirely on FrankaKitchen, underperforming even standard SAC, whereas FMER outperforms SAC-Flow by maintaining precise, closed-form entropy control throughout training.
    \item \textbf{FLAME~\cite{li2026boostingmaximumentropyreinforcement}}. Although FLAME does not incorporate a policy mirror descent trust region in entropy computation. Both methods exploit the tractability of ODE-based flow matching for closed-form entropy computation and employ Hutchinson's estimator to efficiently approximate the vector field divergence. The key differences lie in architecture and action space handling. FLAME primarily targets discretization bias in one-step FM policies, and addresses this through a decoupled actor-critic design: the actor generates actions in a single step for low-latency inference, while the critic evaluates entropy over a multi-step integration grid to preserve regularization accuracy. For bounded action spaces, FLAME enforces boundary constraints directly on the reverse sampling process via a truncated normal distribution. In contrast, FMER does not constrain the sampling process itself. Instead, we formulate the entire flow in an unbounded latent space and derive an explicit analytical correction for the probability density distortion induced by the $\tanh$ transformation into the bounded action space (\cref{prop:divergence_tanh}). This latent-space formulation preserves the theoretical guarantees of standard flow matching while remaining compatible with simulation-free training.
    \item 
    \textbf{RFM~\cite{li2026reverseflowmatchingunified}}. This work is theory-intensive. One major contribution of RFM is to place existing MaxEnt-style diffusion and flow policy methods under a unified framework: the noise-expectation family and the gradient-expectation family. The paper starts from the MaxEnt soft policy improvement target, whose closed-form solution is a Boltzmann distribution over actions. Because this Boltzmann distribution is unnormalized and difficult to sample from directly, the authors propose RFM, which converts the problem into posterior mean estimation. They use self-normalized importance sampling to estimate these posterior means and introduce Langevin Stein control variates to reduce sampling variance. Overall, RFM mainly follows the MaxEnt perspective and builds on the Boltzmann closed-form solution for soft policy improvement. In contrast, our paper incorporates the entropy constraint through a Lagrangian formulation, and our closed-form objective is derived from policy mirror descent.
\end{itemize}

Another line of work estimates policy entropy in closed form without a diffusion or flow backbone. S²AC~\cite{messaoud2024s} represents the policy as a Stein variational gradient descent (SVGD) sampler that transports particles from a learned Gaussian toward the energy-based distribution over Q-values through a fixed number of deterministic, invertible steps, so that the particle density can be tracked with the change-of-variables formula and its entropy approximated to first order using only first-order derivatives and inner products. The follow-up MET-SVGD~\cite{messaoudparticles} shows that this estimator breaks down beyond a few tens of dimensions; it restores the missing term with Hutchinson’s estimator, derives a global invertibility condition, learns the kernel bandwidth and step size end to end, and adds a Metropolis--Hastings correction. FMER shares the goal of an expressive policy with analytic entropy but reaches it differently: \cref{theo:entropy_lf} is the continuous-time change of variables along an ODE whose flow is invertible by construction, so no step-size, invertibility, or acceptance conditions arise, and Hutchinson’s estimator is applied to the divergence of the learned velocity field rather than to the curvature of the target density. Both SVGD methods are full MaxEnt algorithms with entropy inside the soft Bellman target, whereas FMER imposes entropy only as an actor-side constraint, and neither accounts for bounded actions through an analytic density correction.

\subsection{Connections to Policy Mirror Descent Methods}
\label{appendix:connection_pmd}
\begin{itemize}
    \item \textbf{DPMD~\citep{ma2025efficientonlinereinforcementlearning}.} FMER and DPMD share the same ideal nonparametric policy mirror descent target. The distinction lies in how this shared target is realized and optimized. (1) Different regression objects. DPMD uses a diffusion policy, interpreted as a sequence of noise-perturbed energy-based models (EBMs), which can be optimized using reweighted denoising score matching (DSM). Its tractable target relies on a Gaussian corruption kernel \(q_{t\mid 0}(a_t\mid a_0)\), whose conditional score \(\nabla_{a_t} \log q_{t\mid 0}(a_t\mid a_0)\) is available analytically. FMER uses the deterministic OT path \(x_t = (1-t)x_0 + tx_1\). Conditioned on a fixed endpoint pair \((x_0,x_1)\), this path has a Dirac conditional distribution, so DPMD’s Gaussian-kernel DSM objective does not directly apply to our endpoint-conditioned deterministic path. (2) Different finite-sample normalization. In FMER, we approximate \(MZ(s)\) using \(\sum_{l=1}^{M} \exp\!\bigl(A(s,a^{(l)})/\tau\bigr)\) over \(M\) candidates, producing self-normalized weights. DPMD, following the EBM/DSM route, instead uses unnormalized exponential Q-weights, while its running mean-and-variance normalization is an engineering fix for numerical stability. FMER's contribution is therefore a weighted-CFM realization of an existing PMD principle, combined with tractable, $\tanh$-corrected action-space entropy control.
    
    \item \textbf{FPMD~\cite{FPMD}} FPMD is primarily designed to enable one-step flow-policy inference, whereas FMER retains multi-step ODE sampling and focuses on exploration in sparse-reward settings. For policy optimization, FPMD uses unnormalized exponential weights with a fixed temperature, while FMER uses candidate-wise self-normalized weights and adapts the temperature to maintain a target effective sample size. For exploration, FPMD uses scheduled Gaussian action noise, whereas FMER enforces a minimum-entropy constraint using an analytical action-space entropy objective that accounts for the density distortion induced by the $\tanh$ transformation.
\end{itemize}
\subsection{Hutchinson’s trace estimator}
\label{appendix:Hutchinson}
In practice, computing the divergence $\nabla_x \cdot v_\theta(x) = \sum_{i=1}^d \frac{\partial v_{\theta,i}}{\partial x_i}$ in Eq.~\eqref{eq:loss_ent} requires the trace of the Jacobian matrix, which scales as $\mathcal{O}(d^2)$ because calculating each diagonal entry of the Jacobian requires a separate differentiation pass. This quadratic complexity becomes computationally prohibitive, especially for high dimensional action spaces. To maintain efficiency, we employ Hutchinson’s trace estimator \citep{DBLP:conf/iclr/GrathwohlCBSD19, hutchinson1989stochastic}, which provides an unbiased stochastic estimate of the divergence:
\begin{equation}
\nabla_x \cdot v_\theta(x) = \text{Tr}\left(\frac{\partial v_\theta}{\partial x}\right) = \mathbb{E}_{p(\epsilon)}\left[\epsilon^\top \frac{\partial v_\theta}{\partial x}\epsilon\right],
\end{equation}
where $\epsilon \in \mathbb{R}^d$ is a noise vector sampled from a distribution $p(\epsilon)$ with zero mean and unit variance. We use the Rademacher distribution as recommended by \citet{hutchinson1989stochastic}. By leveraging modern automatic differentiation, this reduction to $\mathcal{O}(d)$ per-step complexity allows FMER to scale gracefully to complex high-dimensional environments, such as Humanoid.

\subsection{Remark on action-space entropy}
\label{appendix:H_0aremark}
Note that the Flow Matching process evolves in the unbounded latent space $x$, while the final policy entropy refers to the bounded action space $a$. The initial entropy in the action space, $H_0^a$, differs from the latent Gaussian entropy $H_0^x$ by a constant Jacobian term. The final policy entropy $H_1^a$ is obtained by evolving this base value according to the divergence of the induced vector field:
\begin{align}
    H^a_0 &= H_0^x + \mathbb{E}_{x_0 \sim p_0} \left[ \sum_{i=1}^d \log \left( 1 - \tanh^2(x_{0,i}) \right) \right], \label{eq:H_a0} \\
    H^a_1 &\approx H^a_0 + \mathbb{E}_{t\sim\mathcal{U}[0,1],x_t \sim p_t(\cdot|s)}\left[\nabla_a \cdot v^a(a, t, s)\right].
    \label{eq:H_1}
\end{align}
Since $H_0^x$ and the Jacobian expectation in Eq.~\eqref{eq:H_a0} depend only on the fixed base distribution $p_0$, $H_0^a$ is constant w.r.t. the policy parameters $\theta$. Therefore, it can be omitted from the optimization objective without affecting the gradient.

\subsection{Limitations and future work}
\label{sec:appendix_limitation}
Despite its strong empirical performance, FMER has several limitations that warrant future investigation. First, as shown in Table~\ref{tab:computation}, while FMER is more efficient than diffusion-based baselines, it remains slower than FM-based counterparts such as SAC-Flow. This gap can be attributed to two factors: FMER requires more ODE integration steps (10 vs. 4), and its closed-form entropy computation, while theoretically principled, demands more computational resources than the entropy surrogates employed by SAC-Flow. This presents a tradeoff: FMER's closed-form entropy control yields superior exploration quality, particularly in sparse-reward settings, but at a higher per-step computational cost. Second, while the Hutchinson estimator reduces the divergence computation from $\mathcal{O}(d^2)$ to $\mathcal{O}(d)$, it introduces variance into the entropy gradient estimates, which may affect convergence. We encountered no computational or memory issues in the evaluated action spaces ($d=3$--$17$), but the cost of entropy estimation also depends on the vector-field architecture and batch size. Substantially higher-dimensional actions or long action chunks may increase this cost considerably. Evaluating and improving scalability in these settings remains an important direction for future work. Third, while our closed-form entropy derivation accounts for the $\tanh$ transformation, it assumes the flow operates in a factorized latent space with an isotropic Gaussian base distribution; extending FMER to structured or non-isotropic base distributions would require revisiting the entropy correction in \cref{prop:divergence_tanh}. Finally, all experiments are conducted in simulation, and the sim-to-real gap for flow-based policies in contact-rich manipulation tasks remains an open question. We view these as natural directions for future work rather than fundamental barriers to the approach.

\section{Experiments}

\subsection{Baseline implementation}
\label{appendix:baseline}
We use the following codebases for the baseline implementations. Hyperparameters are set according to the specifications in their respective papers; otherwise, the default values are used.
\begin{itemize}
    \item PPO: \url{https://stable-baselines3.readthedocs.io/en/master/modules/ppo.html}
    \item SAC: \url{https://stable-baselines3.readthedocs.io/en/master/modules/sac.html}
    \item TD3: \url{https://stable-baselines3.readthedocs.io/en/master/modules/td3.html}
    \item DIPO: \url{https://github.com/mahaitongdae/diffusion_policy_online_rl/tree/main}
    \item QVPO: \url{https://github.com/wadx2019/qvpo/tree/main/agent}
    \item DPMD: \url{https://github.com/mahaitongdae/diffusion_policy_online_rl/tree/main}
    \item SAC-Flow: \url{https://github.com/Elessar123/SAC-FLOW/tree/master/from_scratch_code}
    \item FLAME: \url{https://github.com/lzqw/FLAME}
\end{itemize}

\begin{table}[!h]
\small
\centering
\caption{Hyperparameters used in the baselines. $\dagger$ 20 for training/sampling, 1 for test. *4 for critic update, 32 for evaluation, 64 for actor update. \textbf{Note:} SAC-Flow utilizes multiple sub-networks for the actor with different depths and activations.}
\label{tab:hyperparams}
\resizebox{\columnwidth}{!}{%
\begin{tabular}{l|cccccccc}
\hline
\textbf{Hyperparameters} &\textbf{FLAME} & \textbf{DIPO} & \textbf{DPMD} & \textbf{QVPO} & \textbf{SAC-Flow}  & \textbf{SAC} & \textbf{TD3} & \textbf{PPO} \\
\hline
No. of hidden layers &3 & 3 & 3 & 2  &Multiple  & 2 & 2 & 2 \\
No. of hidden nodes   & 256 & 256 & 256 & 256  & Multiple  & 256 & 256 & 256 \\
Activation   &mish & mish & mish & mish &Multiple  & ReLU & ReLU & ReLU \\
Batch size &256 & 256 & 256 & 256 & 512  & 256 & 256 & 256 \\
Discount for reward $\gamma$    & 0.99 & 0.99 & 0.99 & 0.99 &0.99 & 0.99 & 0.99 & 0.99  \\
Reward scale &0.2 &0.2 &0.2 &1 &1  &1 &1 &1\\
Target smoothing coefficient  & 0.005 & 0.005 & 0.005 & 0.005 &N/A  & 0.005 & 0.005 & N/A  \\
Learning rate for actor          &\multicolumn{3}{c}{$3\times 10^{-4}$, annealing to $5\times 10^{-5}$} & $3\times 10^{-4}$ & $3\times 10^{-4}$ & $3\times 10^{-4}$ & $1\times 10^{-3}$& $3\times 10^{-4}$ \\
Learning rate for critic      & $3\times 10^{-4}$  & $3\times 10^{-4}$ & $3\times 10^{-4}$  & $3\times 10^{-4}$ &$1\times 10^{-3}$ &$3\times 10^{-4}$ & $1\times 10^{-3}$ & $3\times 10^{-4}$ \\
Diffusion steps  & $20/1^\dagger$ & 100 & 20 & 20 &4 & N/A & N/A & N/A \\
Candidate set size &32 & 1 & 32 & $4/32/64^*$ &- & N/A & N/A & N/A \\
\hline
\end{tabular}%
}
\end{table}

\subsection{FMER implementation}
\label{appendix:fmer_hp}
\begin{table}[!h]
\small
\centering
\caption{Hyperparameters used in FMER (shared across all environments).}
\begin{tabular}{l|c}
\hline
\textbf{Hyperparameters} & \textbf{Value} \\
\hline
No. of hidden layers            & 3   \\
No. of hidden nodes             & 256   \\
Activation                      & mish \\
Batch size                      & 256 \\
Discount for reward $\gamma$    & 0.99  \\
Target smoothing coefficient & 0.005  \\
Shared learning rate for actor  &$3\times 10^{-4}$, annealing to $5\times 10^{-5}$ \\
Learning rate for critic        & $3\times 10^{-4}$ \\
Advantage weight initialized $\tau$ &0.5 \\
ODE steps    & 10 \\
Target $\mathrm{ESS}^*$ &3\\
Target entropy $\bar{H}$ & Annealed from $0$ to $-dim(\mathcal{A})$\\
Candidate set size & 8\\
Initial $\tau$ value &0.5\\
Initial $\alpha$ value &$e$ \\
Optimizer &Adam \cite{adam} \\
\hline
\end{tabular}%
\end{table}

\subsection{Action generation trajectory in \textit{multi-goal} environment}
\label{appendix:multi_goal}
\begin{figure}[!h]
    \centering
    \includegraphics[width=0.4\linewidth]{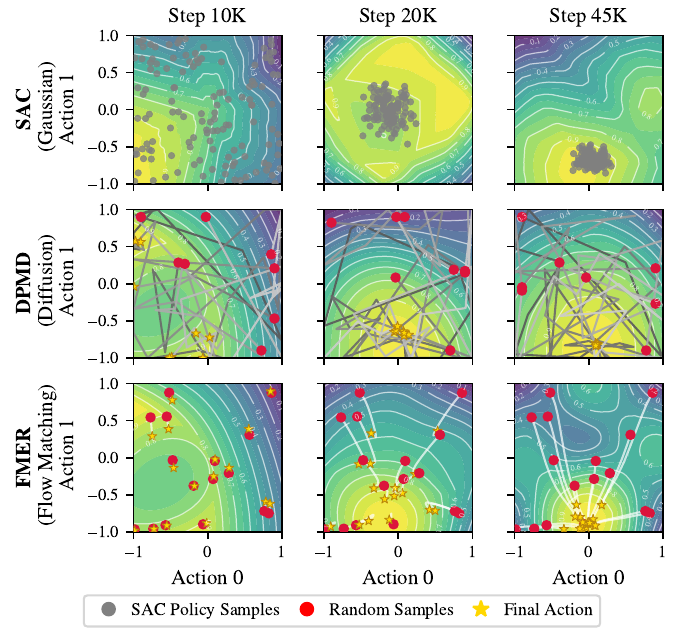}
    \caption{
    Policy evolution on the 2D multi-goal task for the same state \(s=(0,0)\). Background contours depict the learned critic \(Q(s,a)\) over action space. For SAC, points show policy samples, and for the generative policies, red dots show the random noise samples and their sampling trajectory towards the final action.
    }
    \label{fig:train_evolution}
\end{figure}

Figure~\ref{fig:train_evolution} visualizes, for the same state \(s=(0,0)\), the action samples produced by each policy at an early (10K), intermediate (20K), and converged (45K) stage of training. 
SAC initially concentrates samples in a broad, unimodal region and only gradually shifts mass toward a high-value basin, reflecting the limited ability of a single Gaussian to represent disconnected modes. 
In contrast, both DPMD and FMER allocate probability mass to the high-value region much earlier, indicating faster discovery of promising modes. Diffusion denoising trajectories are often indirect and vary across samples, whereas FMER induces smoother and more coherent transport toward high-value actions.

\subsection{FrankaKitchen environment setup}
\label{appendix:frankakitchen_setup}
The environment details are available at \url{https://robotics.farama.org/envs/franka_kitchen/}.
\begin{table}[h]
\caption{FrankaKitchen task description.}
    \centering
    \begin{tabular}{c|l}
        \hline
        \textbf{Tasks} & \textbf{Description} \\
        \hline
        ``top burner"  & Turn the oven knob that activates the top burner\\
        ``bottom burner"& Turn the oven knob that activates the bottom burner\\
         ``light switch" &Turn on the light switch  \\
        ``slide cabinet" &Open the slide cabinet \\
        ``hinge cabinet"  &Open the left hinge cabinet\\
        ``microwave" &Open the microwave door\\
        ``kettle" &Move the kettle to the top left burner\\
         \hline
    \end{tabular}
\end{table}

\begin{table}[!h]
\caption{FrankaKitchen task compositions.}
    \centering
    \resizebox{\linewidth}{!}{%
    \begin{tabular}{c|l}
        \hline
        \textbf{Tasks ($N$)} & \textbf{Composition} \\
        \hline
         $N=1$ & ``light switch" \\
         $N=2$ & ``light switch", ``slide cabinet" \\
         $N=4$ & ``light switch", ``slide cabinet", ``bottom burner", ``microwave" \\
         $N=7$ & ``light switch", ``slide cabinet", ``bottom burner", ``microwave", ``kettle", ``top burner", ``hinge cabinet" \\
         \hline
    \end{tabular}%
    }
\end{table}

\subsection{FrankaKitchen performance}
\label{appendix:franka_perf}
\begin{table*}[!h]
\small
\centering
\caption{We report the mean number of accomplished tasks and the normalized task completion rate (\%). Results reflect the maximum evaluation return achieved during the final 10\% of training (steps 900K to 1M), averaged across 5 random seeds. Each evaluation point is computed over 10 independent environment episodes. Standard deviations are provided in parentheses.}
\resizebox{\linewidth}{!}{%
\begin{tabular}{l|cccc|cccc}
\toprule
&\multicolumn{4}{c|}{\textbf{Accomplished Tasks}} &\multicolumn{4}{c}{\textbf{Normalized Task Completion Rate (\%)}} \\
\textbf{Method} & \textbf{N=1} & \textbf{N=2} & \textbf{N=4} & \textbf{N=7} & \textbf{N=1} & \textbf{N=2} & \textbf{N=4} & \textbf{N=7} \\
\midrule
\textbf{PPO} & 0.02 (0.04) & 0.00 (0.00) & 0.04 (0.09) & 0.60 (0.55) & 2.00 (4.47) & 0.00 (0.00) & 1.00 (2.24) & 8.57 (7.82) \\
\textbf{SAC} & 0.40 (0.55) & 0.64 (0.50) & 0.10 (0.14) & 0.86 (0.30) & 40.00 (54.77) & 32.00 (24.90) & 2.50 (3.54) & 12.29 (4.24) \\
\textbf{TD3} & 0.00 (0.00) & 0.22 (0.49) & 0.60 (0.69) & 0.40 (0.55) & 0.00 (0.00) & 11.00 (24.60) & 15.00 (17.32) & 5.71 (7.82) \\
\textbf{DIPO} & 0.00 (0.00) & 0.66 (0.61) & 0.58 (0.86) & 1.62 (1.29) & 0.00 (0.00) & 33.00 (30.33) & 14.50 (21.39) & 23.14 (18.36)\\
\textbf{DPMD} & 0.60 (0.55) & \textbf{1.00 (0.00)} & 1.00 (0.71) & 1.44 (0.61) & 60.00 (54.77) & \textbf{50.00 (0.00)} & 25.00 (17.68) & 20.57 (8.67)\\
\textbf{QVPO}  & 0.00 (0.00) & 0.00 (0.00) & 0.00 (0.00) & 1.32 (0.66) & 0.00 (0.00) & 0.00 (0.00) & 0.00 (0.00) & 18.86 (9.44)\\
\textbf{SAC-Flow} & 0.00 (0.00) & 0.00 (0.00) & 0.00 (0.00) & 0.04 (0.09)  & 0.00 (0.00) & 0.00 (0.00) & 0.00 (0.00) & 0.57 (1.28)\\
\textbf{FLAME} & 0.08 (0.18) & 0.20 (0.45) & 0.26 (0.58) & 1.28 (0.54) & 8.00 (17.89) & 10.00 (22.36) & 6.50 (14.53) & 18.29 (7.79) \\
\textbf{FPO} & 0.00 (0.00) & 0.02 (0.04) & 0.00 (0.00) & 0.00 (0.00)  & 0.00 (0.00) & 1.00 (2.24) & 0.00 (0.00) & 0.00 (0.00)\\
\midrule
\textbf{FMER} &\textbf{ 0.80 (0.45)} & {0.96 (0.09)} &\textbf{ 1.86 (0.15)} & \textbf{2.82 (1.47)}  & \textbf{80.00 (44.72)} & {48.00 (4.47)} & \textbf{46.50 (3.79)} & \textbf{40.29 (21.03)}\\
\bottomrule
\end{tabular}%
}
\end{table*}

We observe several notable performance trends among the baselines:
\begin{itemize}
    \item Task completion versus normalized task completion rate: Across most baselines, as the number of tasks $N$ increases, the absolute number of completed tasks rises, whereas the overall normalized task completion rate declines. The number of completed tasks rises because the number of total tasks rises. This suggests that agents readily exploit isolated, simpler sub-tasks for initial rewards. However, the fixed episode budget of 280 steps imposes a bottleneck on sequential execution, leaving insufficient time to complete subsequent sub-tasks.
    \item SAC exhibits a non-monotonic pattern across task configurations, with a sharp drop at $N=4$ despite $N=4$ being a superset of the $N=2$ tasks. We hypothesize this reflects the limitation of Gaussian modelling. The Gaussian policy's limited expressiveness becomes particularly damaging at $N=4$, where optimal behavior requires committing to one of several distinct sequential task orderings. A unimodal policy attempts to average over these modes and commits to none. The recovery at $N=7$ is consistent with the partial-reward accumulation effect noted above: with eight available sub-tasks, even a poorly committed unimodal policy can accumulate rewards by accidentally completing simpler sub-tasks encountered during exploration. 
    \item DPMD susceptibility to local optima: DPMD achieves the highest performance among baselines for low $N$. Its KL-divergence constraint acts as a trust region that quickly locks onto a single successful task early in training; however, as $N$ increases, this rigidity hinders exploration and traps the policy in local optima.
    \item FPO performed substantially worse than FMER. A key distinction is the training regime: FPO is on-policy and updates only from recently collected rollout batches, whereas FMER repeatedly reuses replay-buffer transitions with a bootstrapped critic. Under our fixed 1M-step interaction budget, this gives FMER more policy and critic updates per environment transition. This gap is further amplified on FrankaKitchen by the sparse reward structure: successful transitions are rare, and FMER can repeatedly reuse them via its replay buffer.
\end{itemize}
Overall, these findings underscore the necessity of multimodal policy expressiveness and demonstrate the importance of principled exploration control and policy mirror descent.

\subsection{MuJoCo training curves and dynamics}
\label{appendix:mujoco_curves}
\begin{figure}[!h]
    \centering
    \includegraphics[width=0.92\textwidth]{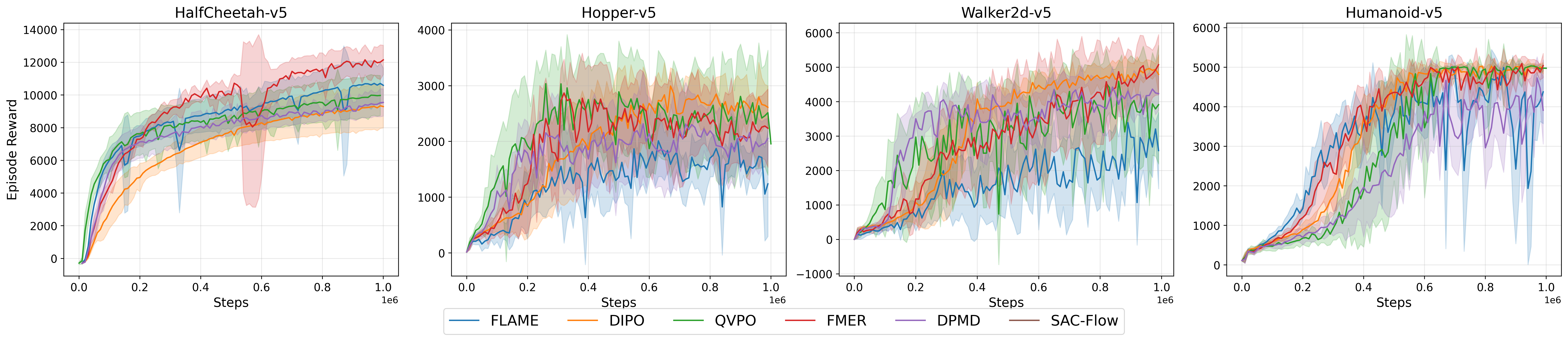}
    \caption{Training curves on four MuJoCo locomotion benchmarks. We report the average episodic return over 5 random seeds, with solid lines representing the mean and shaded regions indicating the standard deviation. For visual clarity, we compare FMER against the five top-performing baselines (QVPO, DPMD, DIPO, FLAME, SAC-Flow). Curves are smoothed with a 10,000-step window.}
    \label{fig:mujoco_curves}
\end{figure}

\noindent\textbf{Training curves.} As shown in Figure~\ref{fig:mujoco_curves}, FMER has a slower convergence rate than other baselines. This is because the Lagrangian multiplier $\alpha$ enforces an entropy constraint to promote early exploration and prevent premature mode collapse. While this sustained exploration enables the discovery of higher-reward solutions in complex spaces, it comes at the cost of more gradual convergence.  On Hopper, it also causes performance to peak and then degrade in simpler, strictly unimodal environments.

\begin{figure}[ht]
    \centering
    \begin{subfigure}[b]{0.24\textwidth}
        \centering
        \includegraphics[width=\textwidth]{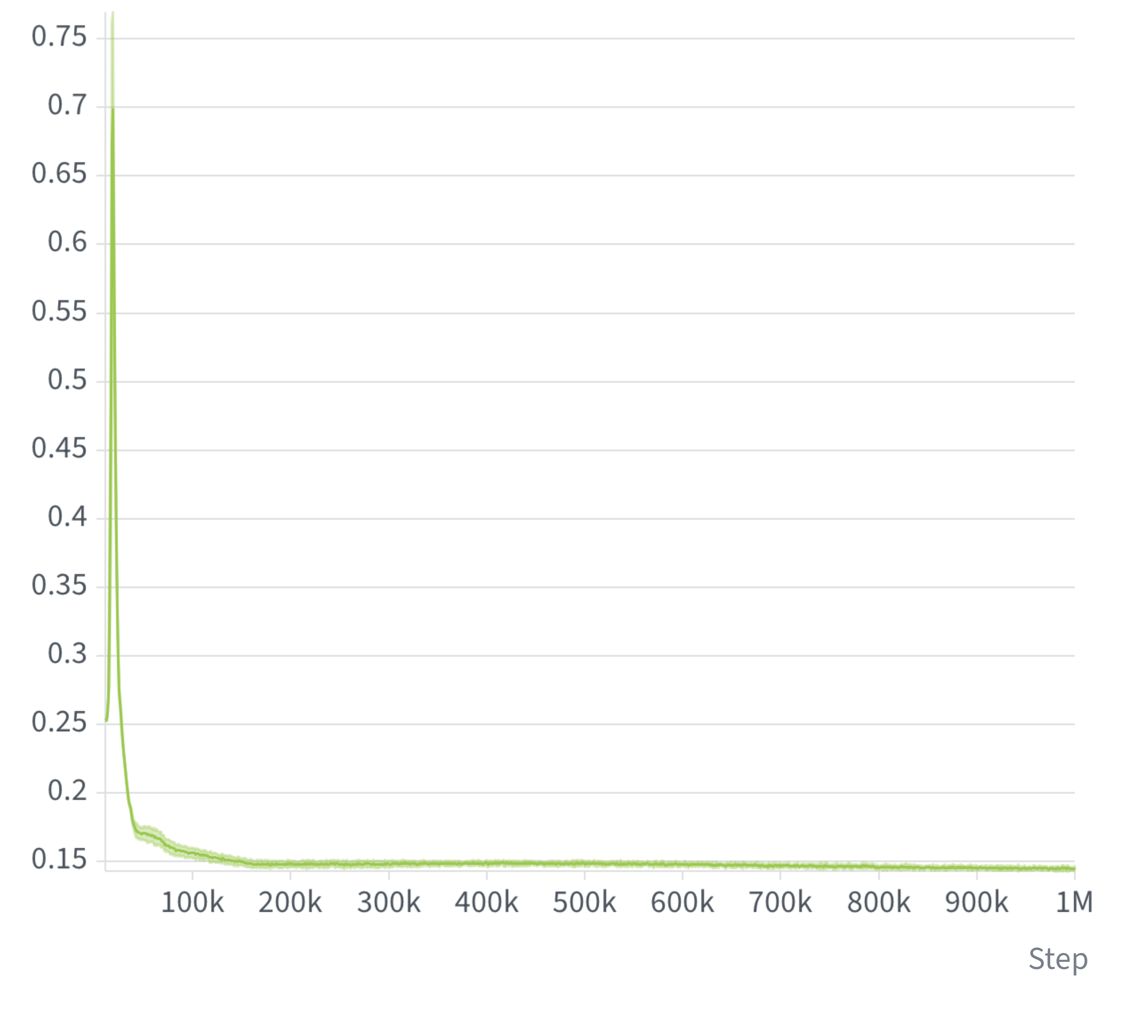}
        \caption{Weighted-CFM loss}
        \label{fig:dyn-cfm}
    \end{subfigure}
    \hfill
    \begin{subfigure}[b]{0.24\textwidth}
        \centering
        \includegraphics[width=\textwidth]{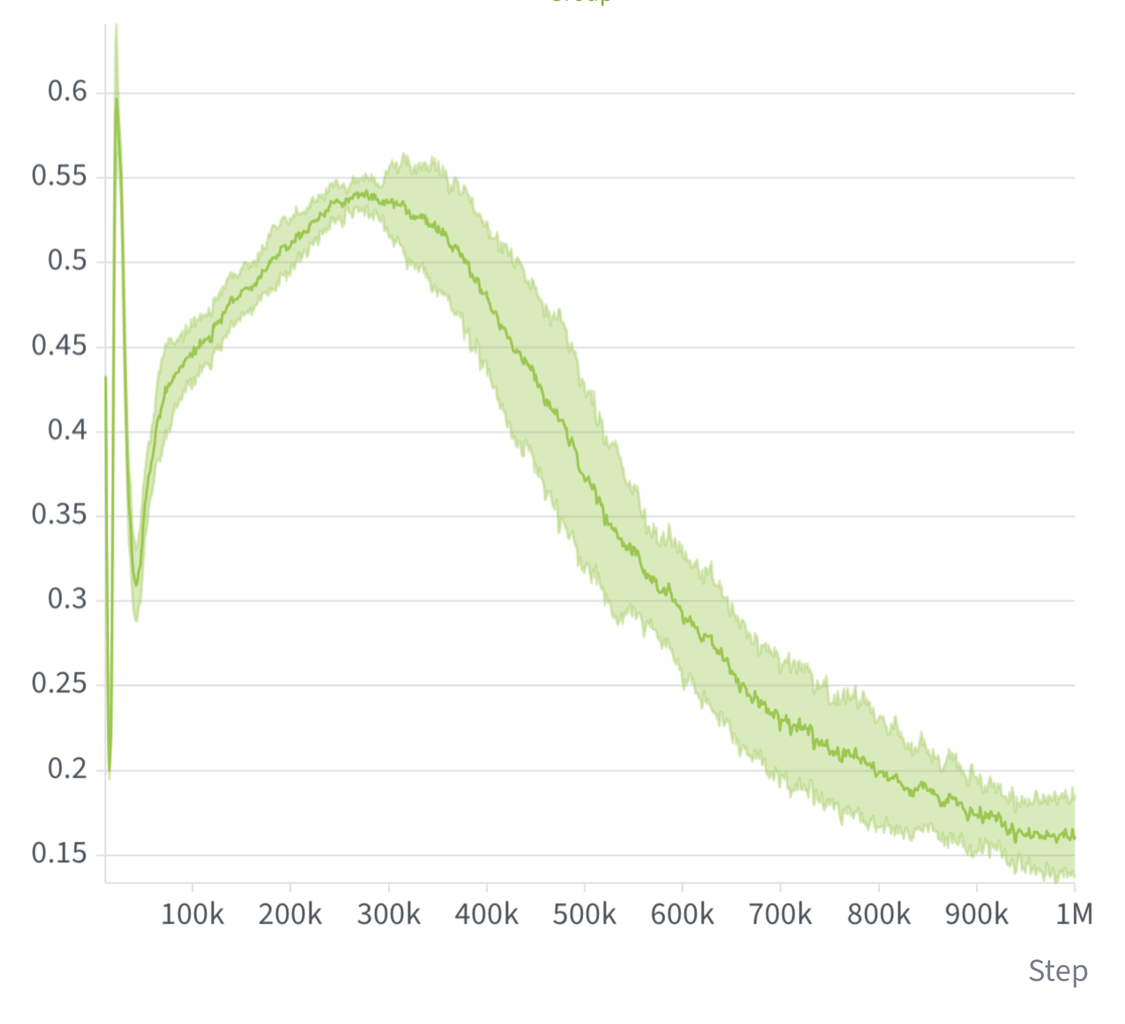}
        \caption{Temperature $\tau$}
        \label{fig:dyn-tau}
    \end{subfigure}
    \hfill
    \begin{subfigure}[b]{0.24\textwidth}
        \centering
        \includegraphics[width=\textwidth]{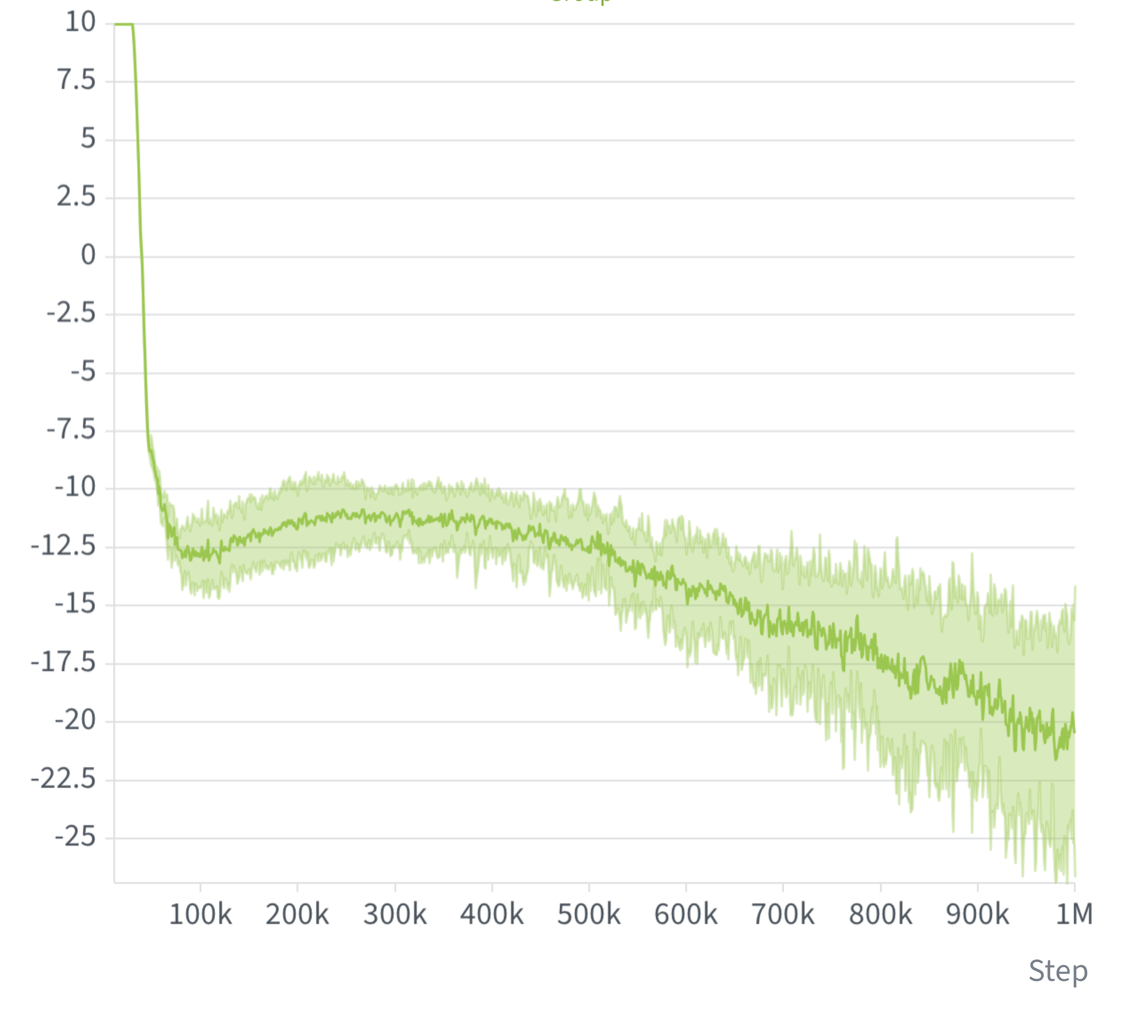}
        \caption{Policy entropy}
        \label{fig:dyn-ent}
    \end{subfigure}
    \hfill
    \begin{subfigure}[b]{0.24\textwidth}
        \centering
        \includegraphics[width=\textwidth]{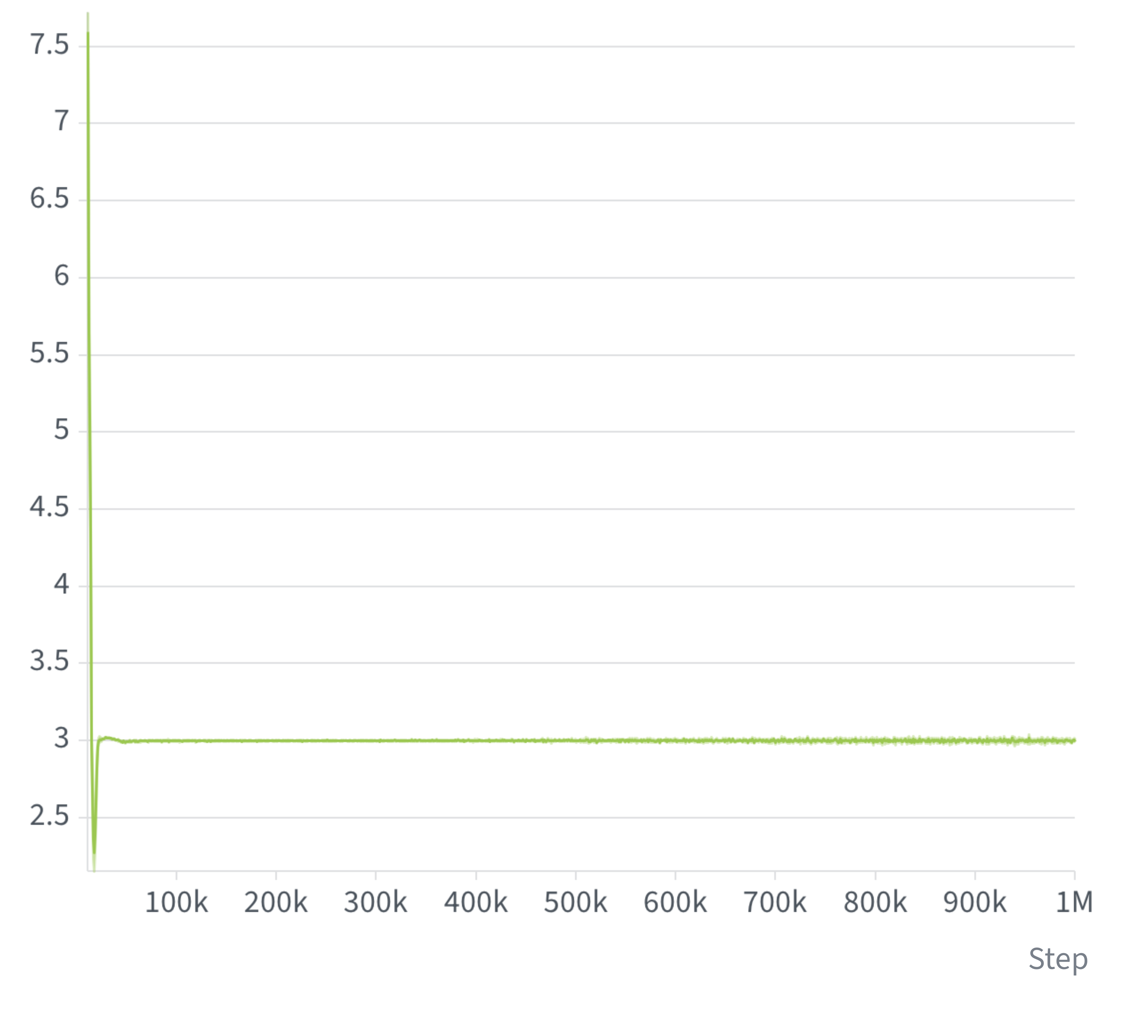}
        \caption{Effective Sample Size}
        \label{fig:dyn-ess}
    \end{subfigure}
    \caption{Training dynamics of FMER on Humanoid-v5, averaged over five seeds; shaded regions denote one standard deviation.}
    \label{fig:training-dynamics}
\end{figure}

\noindent\textbf{Training dynamics.} Figure~\ref{fig:training-dynamics} reports the weighted-CFM loss, temperature $\tau$, policy entropy, and ESS over training on Humanoid-v5, averaged over five seeds. The brief transient at the start reflects the untrained critic: all candidates initially receive similar values, so the weights are uniform and ESS is close to $M$; once the critic begins to discriminate, ESS undershoots and the controller raises $\tau$ to restore the target within a few thousand steps. Thereafter, the weighted-CFM loss decreases monotonically and plateaus early with negligible variance across seeds, indicating stable optimization of the surrogate objective, and the ESS remains at its target ($\mathrm{ESS}^*=3$) for the rest of training, confirming the effectiveness of adaptive $\tau$. Policy entropy drops quickly from its initial near-random level, plateaus while the entropy constraint is binding, and then decreases gradually as the target entropy $\bar{H}$ is annealed. The final entropy remains finite (a deterministic policy would have differential entropy $-\infty$), confirming that the policy stays stochastic while exploration is progressively reduced.

\subsection{Training and inference time}
\label{appendix:training_time}
Table~\ref{tab:computation} reports total training time and inference time on Humanoid-v5 for FMER and all diffusion- and FM-based baselines, with all models trained on a single NVIDIA V100 GPU with 5 CPU cores (25GB system RAM). For QVPO, we report the training time using the official PyTorch implementation. While a JAX port is available in the DPMD repository, our profiling revealed it to be less optimized, resulting in longer training times than the official PyTorch version.
\begin{table}[h]
\small
    \centering    
    \caption{Total training time and inference time of 1 million environment interactions on Humanoid-v5.}
    \begin{tabular}{ccc}
    \toprule
    Algorithm &Training time &Inference time \\
    \midrule
    DIPO (JAX)  &294 min &2.268 ms\\
    DPMD (JAX)   &276 min &1.125 ms\\
    QVPO (PyTorch)\  & 19.7 h &1.284 ms\\
    SAC-Flow (JAX) & \textbf{126 min} &0.483 ms\\
    FMER (JAX)  &215 min &0.515 ms\\
    \bottomrule
    \end{tabular}
    \label{tab:computation}
\end{table}
\begin{table}[h!]
    \centering
    \caption{Training time (minutes) and inference time (ms) across environments with varying action dimensions ($d$), averaged over 5 random seeds. }
    \small 
    \begin{tabular}{cccc}
        \toprule
        \textbf{Environment} &\textbf{Action dimension $d$} &\textbf{Training time} &\textbf{Inference time}\\
        \midrule
        Hopper-v5 & 3 &167 &0.471\\
        HalfCheetah-v5 &6 &168 &0.479\\
        Walker2d-v5 &6 &171 &0.480\\ 
        Humanoid-v5 &17 &215 &0.515\\
        \bottomrule
    \end{tabular}
    \label{tab:comput_d}
\end{table}

As shown in Table~\ref{tab:computation}, FMER achieves substantially lower training time than diffusion-based methods, attributable to its smaller candidate action set and fewer ODE integration steps. SAC-Flow attains the shortest training time overall by further reducing ODE steps and replacing closed-form entropy with a surrogate estimate; however, as shown in Section~\ref{sec:benchmark_franka_mujoco}, this approximation comes at the cost of exploration quality in sparse-reward settings.  

We evaluate the impact of action dimensionality on training time in Table~\ref{tab:comput_d}. As expected, training time scales with $d$ due to the $\mathcal{O}(d)$ complexity of the Hutchinson estimator (Section~\ref{sec:entropy_loss}).
\section{Ablation study}
\subsection{Entropy and advantage-weighting ablations on FrankaKitchen}
\label{appendix:abl_franka_ent_adv}

To evaluate the roles of entropy regularization and advantage weighting in sparse-reward settings, we repeat the corresponding Walker2d ablations on FrankaKitchen with $N=7$. All variants retain the default best-of-$M$ action selection for environment interaction and critic-target computation. Results are reported as mean (standard deviation) across five random seeds.

\noindent\textbf{Entropy regularization.}
We deactivate $\mathcal{L}_{\mathrm{ent}}$ either from the beginning of training or after $5\times10^5$ environment steps. For removal from the beginning, we also evaluate a variant that clips $\tau$ from below to prevent numerical instability.

\begin{table}[ht]
    \centering
    \caption{Entropy ablation on FrankaKitchen ($N=7$). Task completion rates are reported in percentages. NaN denotes a non-finite entropy estimate.}
    \label{tab:entropy_franka}
    \small
    \begin{tabular}{lcc}
        \toprule
        Setting & Task completion rate (\%) & Final entropy \\
        \midrule
        Default FMER & 40.29 (21.03) & -8.71 (1.06) \\
        Entropy removed after $5\times10^5$ steps
            & 40.29 (19.73) & -10.43 (2.36) \\
        Entropy removed from the start
            & 0.00 (0.00) & NaN \\
        Entropy removed from the start, $\tau$ clipped
            & 0.00 (0.00) & -201.36 (11.91) \\
        \bottomrule
    \end{tabular}
\end{table}

As shown in Table~\ref{tab:entropy_franka}, removing entropy regularization from the start leads to failure on this sparse-reward task. During training, ESS remains above its target, driving $\tau$ toward zero. Clipping $\tau$ prevents numerical instability but does not restore performance, and the final entropy remains substantially below that of default FMER. In contrast, removing the constraint after $5\times10^5$ steps leaves the mean task completion rate unchanged despite a decrease in final entropy. These results are consistent with the Walker2d ablation and support the importance of entropy regularization during early exploration.

\noindent\textbf{Advantage weighting.}
We replace the normalized exponential advantage weights with uniform weights, $\hat{w}_j(s)=1/M$, while retaining entropy regularization and all other components. This reduces the actor's weighted-CFM term to unweighted conditional flow matching over the sampled candidate actions.

\begin{table}[ht]
    \centering
    \caption{Advantage-weighting ablation on FrankaKitchen ($N=7$).}
    \label{tab:weighting_franka}
    \small
    \begin{tabular}{lc}
        \toprule
        Setting & Task completion rate (\%) \\
        \midrule
        Default FMER & 40.29 (21.03) \\
        Uniform weights & 33.71 (18.23) \\
        \bottomrule
    \end{tabular}
\end{table}

Uniform weighting yields a lower mean task completion rate than default FMER (Table~\ref{tab:weighting_franka}), consistent with the benefit of emphasizing higher-value candidate actions during policy fitting. 

\subsection{Ablation on $\tanh$ correction in entropy derivation}
\label{appendix:tanh}

A natural alternative to our $\tanh$ correction formulation is to replace the $\tanh$ transformation with simple clipping and compute the entropy directly using \cref{theo:entropy_lf}. We first explain why clipping is theoretically incompatible with our framework, then provide empirical validation.

Our framework requires two properties that clipping violates:
\begin{itemize}
    \item  Non-invertibility: \cref{theo:wcfm} requires mapping actions back to latent space via $x_1=\text{arctanh}(a)$. Clipping is a many-to-one operation that maps all out-of-bound values to the boundary $\pm 1$, making the inversion undefined and collapsing the latent representation.
    \item Non-differentiability: The entropy correction in \cref{prop:divergence_tanh} accounts for the probability density distortion induced by the latent-to-action transformation (Appendix \ref{sec:appendx_A3}). This requires computing the derivative of the transformation, which is undefined at the clipping boundaries ($\pm 1$) with clipping operation.
\end{itemize}

Despite this incompatibility, we ran ablation experiments on MuJoCo to empirically compare $\tanh$ against clipping. In the clipping setup, we bypass the invertible latent mapping and use the environment actions directly as the flow targets during training ($x_1 = a$). Because no $\tanh$ transformation is applied, we omit the density correction from \cref{prop:divergence_tanh} and estimate the uncorrected entropy using the standard vector field divergence (Eq.~\eqref{eq:H_1^x}). During inference, the ODE is integrated to produce an unbounded terminal state $x_1$, which is then hard-clipped to ensure valid environment actions: $a = \text{clip}(x_1, -1, 1)$.

\begin{table}[h]
\centering
\caption{Comparison of clipping and tanh transformations across different environments. Values represent mean returns with standard deviations in parentheses.}
\label{tab:results_comparison}
\begin{tabular}{lcccc}
\toprule
\textbf{Method} & \textbf{Hopper} & \textbf{HalfCheetah} & \textbf{Walker} & \textbf{Humanoid} \\
\midrule
clipping & 2704.55 (627.12) & 7701.50 (598.25) & 4733.82 (696.37) & 4378.76 (2021.24) \\
$\tanh$ & \textbf{2913.45 (642.67)} & \textbf{12332.38 (903.67)} & \textbf{5282.89 (437.57)} & \textbf{5286.12 (65.35)} \\
\bottomrule
\end{tabular}
\label{tab:abl_tanh}
\end{table}

As shown in the Table~\ref{tab:abl_tanh}, $\tanh$ consistently outperforms clipping across environments, validating both our architectural choice and the necessity of the closed-form entropy correction in \cref{prop:divergence_tanh}.

\clearpage
\subsection{Ablation on divergence computation: Hutchinson vs. Exact Jacobian}
\label{appendix:abl_Hutchinson}
\begin{table}[h]
\centering
\caption{Comparison of Hutchinson's estimator and exact trace computation on Walker2d-v5.}
\begin{tabular}{lrr}
\hline
\textbf{Method} & \textbf{Performance} & \textbf{Time (min)} \\ \hline
Hutchinson & 5282.89 (437.57) & 171 \\
Exact & 5316.47 (254.97) & 258 \\ \hline
\end{tabular}
\label{tab:performance_comparison}
\end{table}
To evaluate the impact of the Hutchinson trace estimator, we compare it against an exact Jacobian trace computation on Walker2d-v5 ($d=6$). As shown in Table \ref{tab:performance_comparison}, while the exact calculation reduces the standard deviation across seeds, the mean performance gain is limited, and the exact computation increases training time by nearly 50\% even in this low-dimensional action space. Following \cite{DBLP:conf/iclr/GrathwohlCBSD19}, we adopt the unbiased Hutchinson estimator as it provides the necessary scalability with limited impact on final policy performance.

\subsection{Ablation on adaptive $\tau$}
\label{appendix:adaptive_tau}
Table~\ref{tab:fixed_tau} ablates fixed versus dynamically tuned $\tau$. No single fixed $\tau$ performs consistently and the best fixed value varies across benchmarks. Dynamic tuning achieves competitive performance across all environments without environment-specific tuning, making it a more practical default despite not uniformly dominating any single fixed $\tau$.

\begin{table*}[!h]
\small
\centering
\caption{Sensitivity of FMER to temperature $\tau$: fixed values versus $\mathrm{ESS}$-based dynamic tuning.}
\label{tab:fixed_tau}
\resizebox{\linewidth}{!}{%
\begin{tabular}{l|cccc|cccc}
\toprule
&\multicolumn{4}{c|}{\textbf{FrankaKitchen (finished task)}} &\multicolumn{4}{c}{\textbf{MuJoCo-v5}} \\
\textbf{Method} & \textbf{N=1} & \textbf{N=2} & \textbf{N=4} & \textbf{N=7} & \textbf{Hopper} & \textbf{HalfCheetah} & \textbf{Walker} & \textbf{Humanoid} \\
\midrule
\textbf{$\tau=0.2$} & 0.20 (0.45) & 0.98 (0.67) & 1.00 (0.67) & \textbf{3.10 (1.21)} & 2927.44 (531.97) & 11632.94 (859.00) & 5025.65 (350.93) & \textbf{5399.06 (152.65)}  \\
\textbf{$\tau=0.5$} & 0.60 (0.55) & \textbf{1.42 (0.39)} & 1.36 (0.41) &{2.84 (1.44)} & 2972.15 (344.36) & 10888.09 (1873.16) & \textbf{5504.34 (144.18)} & 5310.64 (87.56) \\
\textbf{$\tau=1.0$} & 0.00 (0.00) & 0.42 (0.58) & 0.56 (0.88) & 1.82 (1.10) & \textbf{3222.88 (289.04)} & 10920.01 (2624.63) & 5430.12 (293.93) & 5228.06 (148.65)\\
\midrule
Dynamic &\textbf{ 0.80 (0.45)} & {0.96 (0.09)} &\textbf{ 1.86 (0.15)} & {2.82 (1.47)} & 2913.45 (642.67) & \textbf{12332.38 (903.67)} & {5282.89 (437.57)} & 5286.12 (65.35)\\
\bottomrule
\end{tabular}%
}
\end{table*}


\subsection{Ablation on candidate size}
\label{appendix:abl_candidate}
Table~\ref{tab:abl_candidate_set} gives performance difference with different candidate size. Training time increases with candidate size (more forward/backward passes). For dense tasks (Walker2d, Humanoid), smaller sets ($M=4,8$) suffice as the value function provides accurate guidance. Sparse tasks (FrankaKitchen) benefit from larger sets ($M=16$) for broader exploration. Excessively large sets ($M=32$) introduce value estimation noise, degrading performance universally. $M=8$ is a robust default setting.
\begin{table}[ht]
    \centering
    \caption{Performance and training time (minutes) across candidate sizes $M$ and environments (5 random seeds). }
    \label{tab:abl_candidate_set}
    \small 
    \begin{tabular}{c|cc|cc|cc|cc}
        \toprule
        & \multicolumn{2}{c|}{\textbf{Walker2d-v5}} & \multicolumn{2}{c|}{\textbf{Humanoid-v5}} & \multicolumn{2}{c|}{\textbf{Franka (N=2)}} & \multicolumn{2}{c}{\textbf{Franka (N=7)}} \\
        \textbf{Candidate Size} & Time & Perf & Time & Perf & Time  & Perf & Time  & Perf \\
        \midrule
        4  &151 & 5321.36  &167& 5243.07 &192 & 1.00  &193 & 2.50\\
        8  &171 & 5282.89  &215 & 5286.12  &220 & 0.96  &217& 2.82 \\
        16  &225 & 4889.52  &285& 5129.35  &275& 1.40  &274 & 2.90\\
        32 &320 & 4657.87  &422& 5048.57 &369&1.24  &370 & 1.96\\
        \bottomrule
    \end{tabular}
\end{table}

\subsection{Ablation on ODE steps and solvers}
\label{appendix:ode_steps_solver}
We investigate the impact of ODE integration steps and solver choices on model performance and training efficiency.

As shown in Table~\ref{tab:abl_ode_steps}, performance remains stable across $N \in [5, 20]$. This range requires fewer steps than most diffusion baselines, maintaining a computational advantage (see Table~\ref{tab:computation}). These results confirm that the straight paths conditioned on Optimal Transport require less precise integration compared to the curved trajectories typical of standard diffusion models.

Table~\ref{tab:abl_ode_solver} demonstrates that higher-order solvers increase the Number of Function Evaluations (NFE) per step while offering limited performance improvements. Applying a 4th-order Runge--Kutta (RK4) solver with $N=2$ underperforms significantly, despite utilizing a similar NFE to the Euler method with $N=10$. This suggests that too few integration steps cause insufficient trajectory coverage regardless of solver order. Euler remains the most efficient default. 
\begin{table}[!ht]
    \centering
    \caption{Training time (minutes) and performance on Walker2d across ODE integration steps ($N$), averaged over 5 random seeds. Performance is presented as Mean (Std).}
    \label{tab:abl_ode_steps}
    \small 
    \begin{tabular}{ccc}
        \toprule
        \textbf{$N$} &Time &Perf \\
        \midrule
        5	&147 &5493.64 (282.53)	\\
        10	&171 &5282.89 (437.57)\\
        20	&226 &5591.64 (153.13)\\
        50	&409 &5186.86 (519.63)\\
        \bottomrule
    \end{tabular}
\end{table}

\begin{table}[!h]
    \centering
    \caption{Training time (minutes) and performance on Walker2d across ODE solvers, averaged over 5 random seeds. Performance is presented as Mean (Std).}
    \label{tab:abl_ode_solver}
    \small 
    \begin{tabular}{ccccc}
        \toprule
        ODE solver &ODE steps &NFE &Time &Perf \\
        \midrule
        Euler &10 &10 &171 &5282.89 (437.57)\\
        Runge–Kutta 2 &10	&20 &219 &5519.10 (395.64)\\
        Runge–Kutta 4 &10	&40&340&5486.24 (184.02)	\\
        Runge–Kutta 2 &5 &10 &168 &5234.53 (172.47)\\
        Runge–Kutta 4 &2 &8 &164 & 3988.16 (2367.62)	\\
        \bottomrule
    \end{tabular}
\end{table}


\newpage
\section*{NeurIPS Paper Checklist}

The checklist is designed to encourage best practices for responsible machine learning research, addressing issues of reproducibility, transparency, research ethics, and societal impact. Do not remove the checklist: {\bf The papers not including the checklist will be desk rejected.} The checklist should follow the references and follow the (optional) supplemental material.  The checklist does NOT count towards the page
limit. 

Please read the checklist guidelines carefully for information on how to answer these questions. For each question in the checklist:
\begin{itemize}
    \item You should answer \answerYes{}, \answerNo{}, or \answerNA{}.
    \item \answerNA{} means either that the question is Not Applicable for that particular paper or the relevant information is Not Available.
    \item Please provide a short (1--2 sentence) justification right after your answer (even for \answerNA). 
\end{itemize}

{\bf The checklist answers are an integral part of your paper submission.} They are visible to the reviewers, area chairs, senior area chairs, and ethics reviewers. You will also be asked to include it (after eventual revisions) with the final version of your paper, and its final version will be published with the paper.

The reviewers of your paper will be asked to use the checklist as one of the factors in their evaluation. While \answerYes{} is generally preferable to \answerNo{}, it is perfectly acceptable to answer \answerNo{} provided a proper justification is given (e.g., error bars are not reported because it would be too computationally expensive'' or ``we were unable to find the license for the dataset we used''). In general, answering \answerNo{} or \answerNA{} is not grounds for rejection. While the questions are phrased in a binary way, we acknowledge that the true answer is often more nuanced, so please just use your best judgment and write a justification to elaborate. All supporting evidence can appear either in the main paper or the supplemental material, provided in appendix. If you answer \answerYes{} to a question, in the justification please point to the section(s) where related material for the question can be found.

IMPORTANT, please:
\begin{itemize}
    \item {\bf Delete this instruction block, but keep the section heading ``NeurIPS Paper Checklist"},
    \item  {\bf Keep the checklist subsection headings, questions/answers and guidelines below.}
    \item {\bf Do not modify the questions and only use the provided macros for your answers}.
\end{itemize}


\begin{enumerate}

\item {\bf Claims}
    \item[] Question: Do the main claims made in the abstract and introduction accurately reflect the paper's contributions and scope?
    \item[] Answer: \answerYes{} 
    \item[] Justification: The claims in abstract and introduction are supported by the proofs in Appendix \ref{sec:appendix_proofs} and experiments in Section \ref{sec:experiments_}.

\item {\bf Limitations}
    \item[] Question: Does the paper discuss the limitations of the work performed by the authors?
    \item[] Answer: \answerYes{} 
    \item[] Justification: Please see Appendix \ref{sec:appendix_limitation}.

\item {\bf Theory assumptions and proofs}
    \item[] Question: For each theoretical result, does the paper provide the full set of assumptions and a complete (and correct) proof?
    \item[] Answer: \answerYes{} 
    \item[] Justification: The proof of each theorem and proposition is provided in Appendix~\ref{sec:appendix_proofs}.

    \item {\bf Experimental result reproducibility}
    \item[] Question: Does the paper fully disclose all the information needed to reproduce the main experimental results of the paper to the extent that it affects the main claims and/or conclusions of the paper (regardless of whether the code and data are provided or not)?
    \item[] Answer: \answerYes{} 
    \item[] Justification: The algorithm is described in Algorithm \ref{alg:fmer} and codes are provided in an anonymous GitHub for reviewers. The GitHub will be made public available upon acceptance. 

\item {\bf Open access to data and code}
    \item[] Question: Does the paper provide open access to the data and code, with sufficient instructions to faithfully reproduce the main experimental results, as described in supplemental material?
    \item[] Answer: \answerYes{} 
    \item[] Justification: This paper has described where to find each environment used in the experiment online. Codes are provided in an anonymous GitHub for reviewers. The GitHub will be made public available upon acceptance. 

\item {\bf Experimental setting/details}
    \item[] Question: Does the paper specify all the training and test details (e.g., data splits, hyperparameters, how they were chosen, type of optimizer) necessary to understand the results?
    \item[] Answer: \answerYes{} 
    \item[] Justification: Please see Section~\ref{sec:experiments_}.

\item {\bf Experiment statistical significance}
    \item[] Question: Does the paper report error bars suitably and correctly defined or other appropriate information about the statistical significance of the experiments?
    \item[] Answer: \answerYes{} 
    \item[] Justification: Experiments are executed with 5 random seeds. For each random seed, 10 random runs are used for evaluation. Results are reported with standard deviation.

\item {\bf Experiments compute resources}
    \item[] Question: For each experiment, does the paper provide sufficient information on the computer resources (type of compute workers, memory, time of execution) needed to reproduce the experiments?
    \item[] Answer: \answerYes{} 
    \item[] Justification: Please see Appendix~\ref{appendix:training_time}.
    
\item {\bf Code of ethics}
    \item[] Question: Does the research conducted in the paper conform, in every respect, with the NeurIPS Code of Ethics \url{https://neurips.cc/public/EthicsGuidelines}?
    \item[] Answer: \answerYes{}  
    \item[] Justification: Our research adheres to the code of ethics by using only synthetic, non-personal data from a simulated environment, posing no risk to human subjects, privacy, or societal harm, and we commit to open-sourcing our code and models with proper documentation to ensure reproducibility and responsible use.

\item {\bf Broader impacts}
    \item[] Question: Does the paper discuss both potential positive societal impacts and negative societal impacts of the work performed?
    \item[] Answer: \answerNo{}
    \item[] Justification: This paper presents work whose goal is to advance the field of reinforcement learning. Experiments are conducted on robot controlling in well-controlled simple tasks in simulated environments. 
    
\item {\bf Safeguards}
    \item[] Question: Does the paper describe safeguards that have been put in place for responsible release of data or models that have a high risk for misuse (e.g., pre-trained language models, image generators, or scraped datasets)?
    \item[] Answer: \answerNA{} 
    \item[] Justification: The paper does not pose any such risks.

\item {\bf Licenses for existing assets}
    \item[] Question: Are the creators or original owners of assets (e.g., code, data, models), used in the paper, properly credited and are the license and terms of use explicitly mentioned and properly respected?
    \item[] Answer: \answerYes{} 
    \item[] Justification: We cite the environment used in the experiments in Section \ref{sec:experiments_}.

\item {\bf New assets}
    \item[] Question: Are new assets introduced in the paper well documented and is the documentation provided alongside the assets?
    \item[] Answer: \answerYes{} 
    \item[] Justification: The codebase and implementation details are shared alongside the paper.

\item {\bf Crowdsourcing and research with human subjects}
    \item[] Question: For crowdsourcing experiments and research with human subjects, does the paper include the full text of instructions given to participants and screenshots, if applicable, as well as details about compensation (if any)? 
    \item[] Answer: \answerNA{} 
    \item[] Justification: The paper does not involve experiments with human participants.

\item {\bf Institutional review board (IRB) approvals or equivalent for research with human subjects}
    \item[] Question: Does the paper describe potential risks incurred by study participants, whether such risks were disclosed to the subjects, and whether Institutional Review Board (IRB) approvals (or an equivalent approval/review based on the requirements of your country or institution) were obtained?
    \item[] Answer: \answerNA{} 
    \item[] Justification: The paper does not involve experiments with human participants.

\item {\bf Declaration of LLM usage}
    \item[] Question: Does the paper describe the usage of LLMs if it is an important, original, or non-standard component of the core methods in this research? Note that if the LLM is used only for writing, editing, or formatting purposes and does \emph{not} impact the core methodology, scientific rigor, or originality of the research, declaration is not required.
    \item[] Answer: \answerNA{} 
    \item[] Justification:  The core method development in this research does not involve LLMs as any important, original, or non-standard components.

\end{enumerate}

\end{document}